\documentclass[journal]{IEEEtran}
\usepackage[utf8]{inputenc}
\usepackage{array}
\usepackage{amsmath}
\usepackage{amssymb}
\usepackage[T1]{fontenc}
\usepackage{multicol}
\usepackage{textcomp}
\usepackage{filecontents,lipsum}
\usepackage[noadjust]{cite}
\usepackage{cprotect}
\usepackage{fvextra}
\usepackage[dvipsnames]{xcolor}
\usepackage{graphicx}
\usepackage{subfig}
\usepackage[ruled,vlined]{algorithm2e}
\usepackage{multicol,tabularx,capt-of}
\usepackage{multirow}
\usepackage{float}
\usepackage{stackengine}
\usepackage[hidelinks]{hyperref}
\usepackage{hyperref}
\usepackage{accents} 
\usepackage[font=footnotesize]{caption}
\usepackage{colortbl}
\usepackage{scalerel}
\usepackage{tikz}
\usepackage{xcolor,cite,etoolbox}
\makeatletter 
\pretocmd\@bibitem{\color{black}\csname keycolor#1\endcsname}{}{\fail}
\newcommand\citecolor[1]{\@namedef{keycolor#1}{\color{blue}}}
\makeatother

\usetikzlibrary{svg.path}
\definecolor{orcidlogocol}{HTML}{A6CE39}
\tikzset{
  orcidlogo/.pic={
    \fill[orcidlogocol] svg{M256,128c0,70.7-57.3,128-128,128C57.3,256,0,198.7,0,128C0,57.3,57.3,0,128,0C198.7,0,256,57.3,256,128z};
    \fill[white] svg{M86.3,186.2H70.9V79.1h15.4v48.4V186.2z}
                 svg{M108.9,79.1h41.6c39.6,0,57,28.3,57,53.6c0,27.5-21.5,53.6-56.8,53.6h-41.8V79.1z M124.3,172.4h24.5c34.9,0,42.9-26.5,42.9-39.7c0-21.5-13.7-39.7-43.7-39.7h-23.7V172.4z}
                 svg{M88.7,56.8c0,5.5-4.5,10.1-10.1,10.1c-5.6,0-10.1-4.6-10.1-10.1c0-5.6,4.5-10.1,10.1-10.1C84.2,46.7,88.7,51.3,88.7,56.8z};
  }
}
\newcommand\orcidicon[1]{\href{https://orcid.org/#1}{\scalerel*{
\begin{tikzpicture}
\pic{orcidlogo};
\end{tikzpicture}
}{|}}}

\begin{document}

\title{Rotation Equivariant Feature Image Pyramid Network for Object Detection in Optical Remote Sensing Imagery}

\author{Pourya~Shamsolmoali, \IEEEmembership{Member, IEEE,} Masoumeh~Zareapoor,
            Jocelyn~Chanussot, \IEEEmembership{Fellow, IEEE,} Huiyu~Zhou, and~Jie~Yang
             \thanks{Manuscript received January 26, 2021; revised May 23, 2021; accepted September 1, 2021. This research is partly supported by NSFC, China (No: 61876107, U1803261)and Committee of Science and Technology, Shanghai, China (No. 19510711200). (Corresponding author: Masoumeh Zareapoor.)

P.~Shamsolmoali, M.~Zareapoor, J.~Yang are with the Institute of Image Processing and Pattern Recognition, Shanghai Jiao Tong University, Shanghai 200240, China. Emails: (pshams, jieyang)@sjtu.edu.cn, mzarea222@gmail.com.}
\thanks{J. Chanussot is with the GIPSA-lab, Université Grenoble Alpes, CNRS, Grenoble INP, 38000 Grenoble, France, and also with the Faculty of Electrical and Computer Engineering, University of Iceland, 101 Reykjavik, Iceland (Email: jocelyn.chanussot@grenoble-inp.fr).}
\thanks{H. Zhou is with the School of Informatics, University of Leicester, Leicester LE1 7RH, United Kingdom. (Email: hz143@leicester.ac.uk).}
}
\markboth{IEEE Transactions on GEOSCIENCE AND REMOTE SENSING, VOL. XX, NO. XX}%
{Shamsolmoali \MakeLowercase{\textit{et al.}}}

\maketitle

\begin{abstract}
\textcolor{blue}{To read the paper please go to IEEE Transactions on Geoscience and Remote Sensing on IEEE Xplore.} Detection of objects is extremely important in various aerial vision-based applications. Over the last few years, the methods based on convolution neural networks have made substantial progress. However, because of the large variety of object scales, densities, and arbitrary orientations, the current detectors struggle with the extraction of semantically strong features for small-scale objects by a predefined convolution kernel. To address this problem, we propose the rotation equivariant feature image pyramid network (REFIPN), an image pyramid network based on rotation equivariance convolution. The proposed model adopts single-shot detector in parallel with a lightweight image pyramid module to extract representative features and generate regions of interest in an optimization approach. The proposed network extracts feature in a wide range of scales and orientations by using novel convolution filters. These features are used to generate vector fields and determine the weight and angle of the highest-scoring orientation for all spatial locations on an image. By this approach, the performance for small-sized object detection is enhanced without sacrificing the performance for large-sized object detection. The performance of the proposed model is validated on two commonly used aerial benchmarks and the results show our proposed model can achieve state-of-the-art performance with satisfactory efficiency.
\end{abstract}


\IEEEpeerreviewmaketitle

\section{Introduction}
\IEEEPARstart {O}{bject} detection in RSIs is a substantial and challenging problem and demanded by a large number of applications such as land planning, crop monitoring, military reconnaissance, and urban monitoring. Over the past several years, a significant number of studies have been focused on object detection in RSIs \cite{shi2013ship, li2018hsf}, which particularly rely on handcrafted features or the statistical distributions of objects \cite{wu2019orsim}. These methods have achieved promising results but lack sufficient robustness in different challenging circumstances. On the other hand, Convolution Neural Network (CNN) approaches have been widely used for RSI object detection and classification \cite{qiu2017automatic, xu2020hierarchical, hong2019learnable, hong2018augmented, zhao2020joint, zhang2018feature} in recent years due to their performance. However, object detection faces a number of significant challenges in RSIs. Since objects in RSI are different from those of objects in natural scenes, for example, the distance between the remote sensor and the objects on the ground is changing, the objects on the ground have various orientations, and the background is highly complex. We cannot address these challenges using the existing object detection models. 

Mainly, object detection methods have two categories: single-stage \cite{liu2016ssd, cao2019triply} and two-stage \cite{girshick2015fast, ren2016faster, he2015spatial}  detectors. Single-stage methods as a direct approach regress the default anchors toward detection of bounding boxes by scanning grids on the image. But, in two-stage approaches, first, object proposals are created, then regressed and classified. In general, single-stage approaches are computationally efficient, but have lower detection performance in comparison with two-stage methods \cite{huang2017speed}. The Single Shot Multibox Detector (SSD) \cite{liu2016ssd} is one of the single-stage approaches which recently has demonstrated a promising tradeoff between detection accuracy and efficiency. In SSD, layers with different resolutions perform predictions, in which the earlier layers contribute in prediction of tiny objects while the deeper or last layers are participating in large objects detection. In spite of its detection achievement, SSD struggles to deal with multi-scale object instances. Especially, small objects detection of SSD is not satisfactory \cite{huang2017speed}, because of the poor differential information in shallower layers.

On the other hand, CNNs have achieved substantial improvement on object detection in RSIs \cite{xu2020hierarchical, feng2020progressive, wu2019orsim, wang2019fmssd}. The achievement of CNNs is mainly due to the weights sharing and the transformation equivariant essence of the convolution procedure. Therefore, any process that preserves the neighborhood pixels such as convolution is transformation equivariant. One of the significant results of learning via the weights of convolution is dramatic decrease in the parameters' range. In absence of the transformation invariance theory, different local windows may have different weights. To decrease the amount of learnable parameters in accordance with the number of pixels in an image, weight sharing is proposed \cite{jaderberg2015spatial}.

\begin{figure}
\footnotesize{
  \centering
  \includegraphics[width=0.48\textwidth]{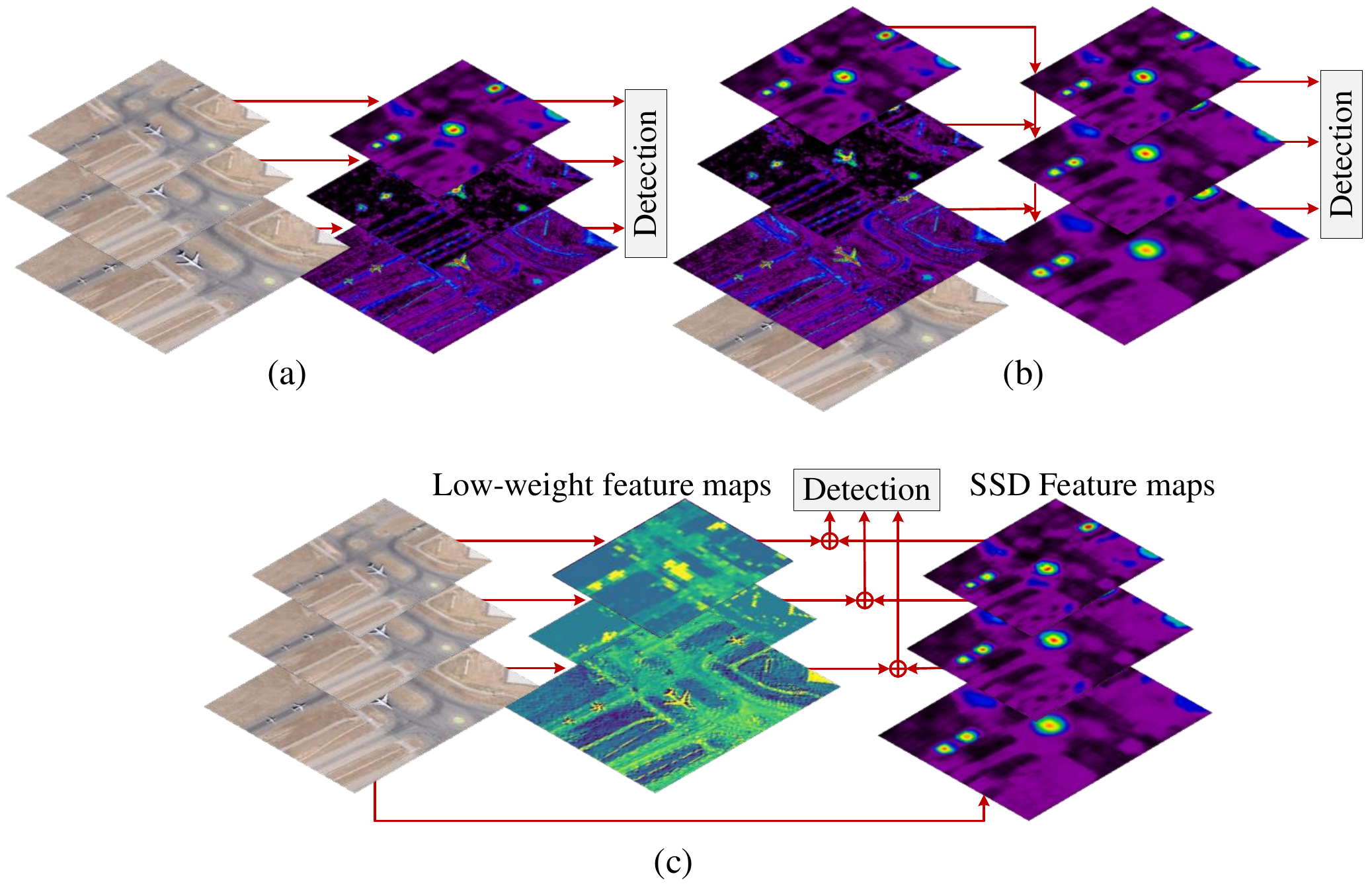}
\caption{Object detection on (a) image pyramid network, (b) feature pyramid network with the top-down architectures, and (c) feature image pyramid network that has strong semantic and fine feature maps.}
\label{fig:1}
}
\end{figure}

Despite the success of CNN models on object detection, learning visual feature representations is a major problem, and detecting multi-scale objects is challenging. To overcome this problem, pyramidal feature representations have been introduced to represent an image through multi-scale features that can be used in object detectors \cite{lin2017feature}.
Feature Pyramid Network (FPN) \cite{lin2017feature} (see Fig. \ref{fig:1}(b)) is one of the best representative approaches for producing pyramidal feature representations of objects. Typically, pyramid models adopt a backbone network and create feature pyramids by successively merging two or three consecutive layers in the backbone network with top-down and adjacent connections. In the standard FPN, the high-level features have lower resolutions but are semantically strong, which can be upscaled and merged with higher resolution features to create stronger representative features. Such network architectures are simple and efficient, but do not have satisfactory performance on detecting small and multi-scale objects with dense distributions \cite{huang2017speed, shamsolmoali2020road} or when the object’s absolute value of orientation in RSIs is not a discriminant feature, due to limited information extracted from previous or shallower layers. \\
To address these issues, this paper proposes a feature image pyramid network that naturally encodes rotation equivariance (REFIPN) to generate semantically strong scale-wise feature maps for object detection in RSIs. In this model, we introduce a feature-merged framework that enhances the context information to improve the small-scales object detection accuracy. More specificly, we improved context information of the feature maps by combining SSD \cite{liu2016ssd} with the proposed light-weight image pyramid module. In our network, a spatial attention module and a feature fusion module are designed to normalise the weights and enrich the context information. In addition, we propose filter CNNs which equivariance under various rotations to obtain rotation equivariance of CNNs with extra weight sharing. The action of rotation equivariance depends on each data sample, with the adequate behaviour learned during training. REFIPN contains superior characteristics when dealing with rotations: by encoding rotation equivariance in the model, we can massively reduce the range of parameters while improving the performance. It is worthwhile to mention, employing the logic of weight sharing for rotations is not a simple task. By heeding this rationale, we apply $\Lambda$ rotations for each single convolution filter, which leads to $\Lambda$ additional feature maps for each filter. In this configuration the rotation weight sharing is given by the manipulation part of the weights which are different from the angle. In this process, each convolution layer do not have any knowledge about the features' orientation of the earlier layers. 

To minimize the model's size while maintaining its robustness against rotations, we only propagate the highest amplitude values obtained from $\Lambda$ feature maps. In the REFIPN (see Fig. \ref{fig:1}(c)), we first down-sample an input image to build an image pyramid module, in which all the stages of the network is featurized. Next, scale-wise features of the pyramidal module and region of interests are integrated into the SSD features (SSD adopts convolution filters), in a spatial attention module, with the aim of increasing the discriminative ability. Also, we utilize a feature integration module to combine the features from different layers. Each layer of the proposed model only keeps the highest amplitude value of $\Lambda$ which appears as a 2D vector field to propagate it to the other layers in the network.
The main contributions of our work are threefold.
\begin{itemize}
\item{We propose a novel object detection model by introducing a feature image pyramid network and integrating multi-level features to obtain more semantically strong features and improve the accuracy of detection.}
\item{We transform the convolution operator by encoding rotation equivariance and shift to deal with the complicated object deformation in RSIs.}
\item{We conducted a comprehensive ablation study to demonstrate the impact of each proposed module on detection results.}
\end{itemize}

The remainder of this paper is organized as follows. Section \ref{sec:2} presents the related work. Section \ref{sec:3} describes the proposed framework, including feature image pyramid network and rotation equivariance that is encoded in the convolution, and feature learning. The experimental results on two datasets are reported in Section \ref{sec:4} to validate the performance of the proposed model. Section \ref{sec:5} concludes this paper.

\section{RELATED WORK}
\label{sec:2}
In the last few years, due to advances of deep learning, there has been considerable advancements in object detection. The object detection methods are generally have two categories: two-stage and one-stage detectors. Two-stage approaches generally use a number of different scale boxes for the input image and then perform detection using standard classifiers. In R-CNN \cite{girshick2014rich}, the model first produces category-independent region proposals, and uses a CNN model to extract a feature vector from the proposed regions, and then predicts the corresponding category by using support vector machines. The R-CNN has a high detection rate, but its speed is limited. With the intention of increasing the speed and accuracy of detection, Fast R-CNN \cite{girshick2015fast} is proposed to use bounding-box regression with an efficient training process. Later, In \cite{ren2016faster}, Faster R-CNN is proposed by combining object proposals and detection into a single unified network that has better efficiency. This model generates a pyramid of feature maps. It applies the region proposal network (RPN) to generate region of interests (ROIs). Several single-stage detection methods have also been proposed, including YOLO \cite{ redmon2016you, redmon2017yolo9000} and SSD \cite{liu2016ssd}. As an example, RetinaNet \cite{lin2017focal} shows better performance as compared to two-stage detection methods while maintaining efficiency. Despite such advances, these approaches do not have satisfactory detection performance in RSIs because of the bird's-eye-view.

\subsection{Object Detection in Aerial Images}
In comparison with detection of an object in natural scene images, object detection in RSIs has additional challenges. This topic has been intensively studied over the last ten years. Conventional object detection models learn to classify the sliding windows or parts of objects to categorize objects and background \cite{qiu2017automatic}. In \cite{li2018hsf}, a feature-based method is proposed for ship detection in RSIs. This method, to detect multi-scale ships, uses a selective detector that generates candidates from the extracted features and introduced a discriminative method to map features from various scales to the same scale for better detection. In \cite {niu2021hybrid} a novel attention network is proposed to capture long-range dependencies and reconstruct the feature maps for better representation in RSIs. 
In \cite{dong2019sig}, a transfer learning model is proposed by adopting Faster R-CNN \cite{ren2016faster}. The proposed model changes conventional non-maximum suppression in the network and minimizes the possibility of missing small objects. Moreover, transfer learning has been used to support RSIs by annotating both object positions and classes. \\
In \cite{feng2020progressive, shamsolmoali2021multi}, weakly supervised object detection methods in RSIs are proposed. The authors designed contextual instance refinement models that have significant attention to diverse objects and object parts. In \cite{sun2021pbnet}, the authors proposed a unified part-based CNN, which is especially developed for composite object detection in RSIs. This model treats a composite object as a set of parts and integrates part details into context information to boost object detection. In \cite{zheng2020foreground}, to solve the problem of foreground-background imbalance in large scale RSIs, the authors proposed a foreground-aware model to detect objects from a complex background. Moreover, in \cite {zheng2020hynet}, a hyper-scale object detection framework is proposed that learns from hyper-scale feature representation to ease the scale-variation problem. Feature fusion is a key technique to enhance the quality of extracted features for object detection. In \cite{hong2021interpretable, hong2020more, hong2020graph, liu2018path} different feature fusion strategies and nonconvex modeling are studied for a better feature representation in RSIs. However, the above methods predict attributes of a bounding box independently and require a strategy for more accurate predictions.

In early studies, for better representation, hand-crafted features were adopted. For example, in \cite{wu2019fourier, guo2020rotational}, the authors are proposed an efficient detection framework based on rotation-invariant feature aggregation and adopted a learning based approach to extract high-level or semantically meaningful features for object detection. Recently, based on \cite { ma2018arbitrary}, several arbitrary-oriented detectors for object detection in RSIs are developed. In \cite{cheng2016learning}, the authors proposed an invariant CNN for target detection in RSIs. Compared to previous models that only optimized by a structural regression, this model is trained by adopting a regularization strategy, which uses the representive features of the training samples before and after rotation. In \cite{li2017rotation}, a multi-scale object detection model in RSIs is proposed to use a double-channel feature aggregation path to learn both local and contextual features. This model has significant detection performance on multi-scale objects but has less efficiency due to the multi-path connections. \par
In \cite {fu2020point, fu2020rotation}, the authors proposed novel methods based on point sampling to improve oriented object recognition and localization. In these models to utilize the spatial information, the detectors encode an oriented object with a point-based representation and operate fully convolution networks for point localization. In these models to enhance localization accuracy, the detectors use the coarse-to-fine method to reduce the quantization error in point localization. Ding et al. \cite {ding2019learning} proposed a model named RoITransformer by activating spatial transformations on ROIs and learn the transformation parameters under the supervision of oriented bounding box annotations. RoITransformer has lightweight and can be adopted by other detectors for oriented object detection. In \cite{wei2020x}, an aircraft detection model is proposed based on prediction and clustering of paired intersecting line segments of each object. This model operates based on line segments estimation, without performing rectangular regions classification or using features learning. In \cite{xu2020hierarchical}, a hierarchical semantic framework is proposed to improve object detection performance in RSIs. This model integrates spatial attention details and the global semantic information to obtain more discriminative features. However, these works use axis-aligned anchors, and sampling point methods to localize and detect objects, but, the misalignment between boxes and candidate objects prevent accurate detection of small objects. Moreover, these models are complicated due to complex feature extraction and need a lot of GPU resources. 

The distinct imaging conditions and varied section patterns bring untreated challenges to object detection in RSIs. Consequently, it is hard to obtain satisfactory results by directly deploying the available object detection models. Furthermore, due to the deficit of training data and the complexity of the network to handle various objects with multiple scales and confusing background, the excellence of deep learning has not been signified in object detection of RSIs. To respond to the above problems, we introduce a novel method for object detection in RSIs. Different from the existing methods, our REFIPN only utilize and incorporate hierarchical semantic information that are extracted by our optimized model which consist of light-weight image pyramid module that integrated into SSD module. Moreover, a novel convolution filter is introduced for better orientation detection.
\begin{figure*}
  \centering
  \includegraphics[width=0.86\textwidth]{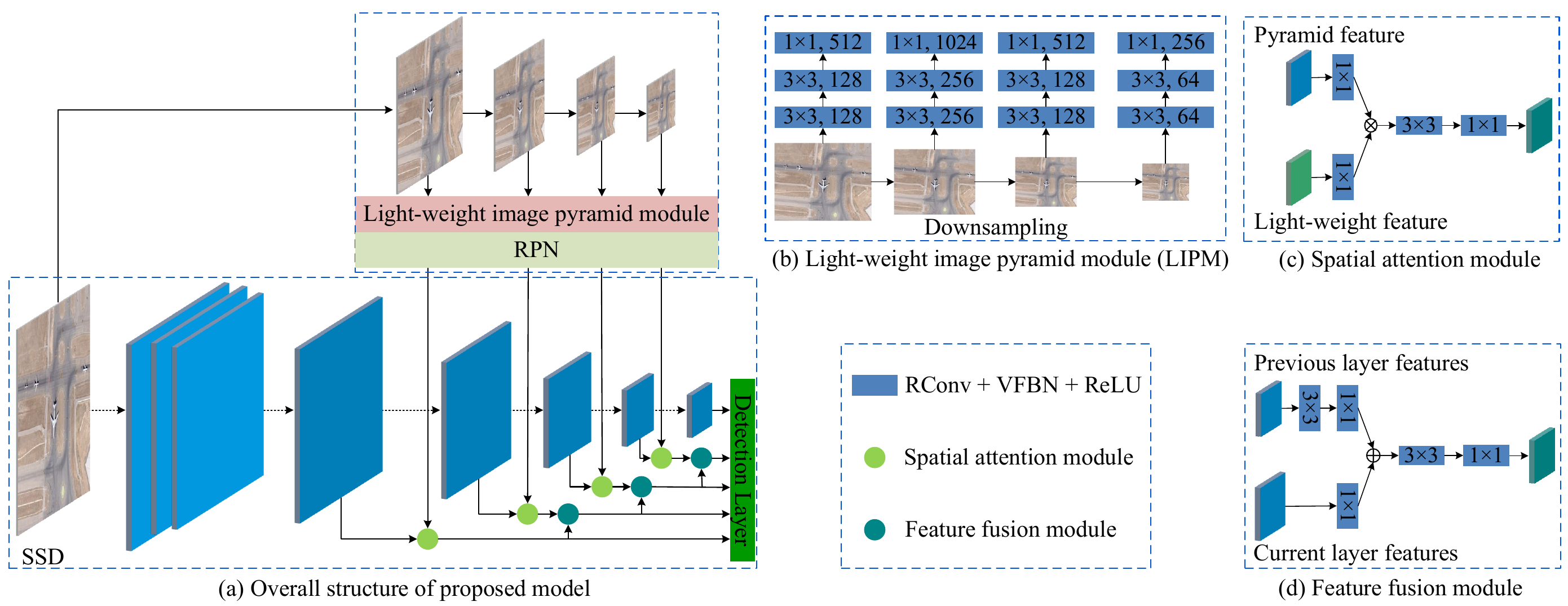}
\caption{\footnotesize Architecture of our proposed detector. (a) Basic structure. The proposed model uses SSD as the baseline and extend it with a light-weight image pyramid module to project ROI at different scales to feature map, details are illustrated in (b). In (c) and (d), we have shown the architectures of the spatial attention and the feature fusion modules respectively.}
\label{fig:2}
\end{figure*}
\begin{figure}
  \centering
  \includegraphics[width=0.5\textwidth]{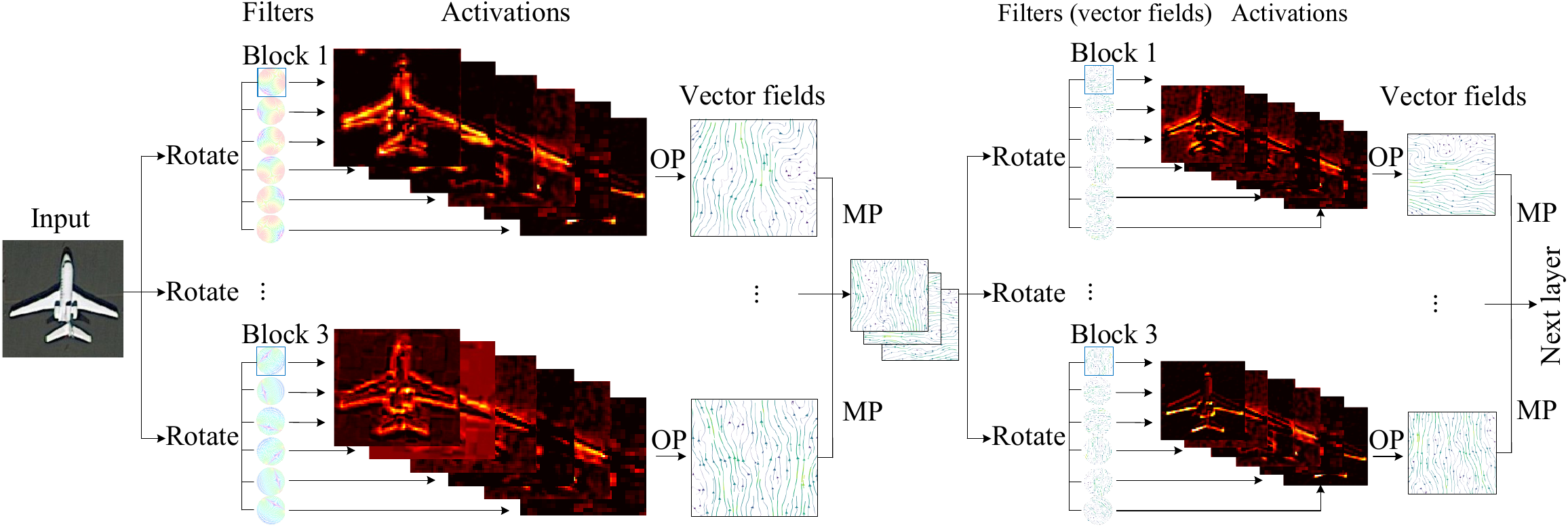}
\caption{\footnotesize Processing details of the RConv. Each single layer just learns three canonical filters (blue rectangles) and reproduces them over six orientations. Three vector field maps are obtained from the first block, further in the next block, the vector field maps are convolved by vector field filters (OP and MP denote orientation pooling and max pooling respectively).}
\label{fig:3}
\end{figure}\\

\subsection{Filters Rotation for Convolution }
Equivariance to translations and deformations are firstly studied in \cite{lecun2015deep}. To perform convolution on an input image $x\in \mathbb{R}^{M\times N\times c}$ alongside a filter $f\in \mathbb{R}^{m\times n\times c}$ ($M\times N$ denotes the image size and $c$ is channels), the output $y=f*x$ is obtained by utilizing the same operator for all the projected $m\times n$ sliding windows on $x$. If $x$ engages in integer transformation in the length and breadth orders by ($u, v$) pixels, the same neighbouring pixel in $x$ will be presented in the transformed $x$, and needs to be transformed by ($u, v$) pixels. One of the main equivariance to translations models is Spatial Transformer Network (STN) \cite{jaderberg2015spatial} which learns a canonical pose and produces an invariant representation through warping. In \cite{esteves2017polar}, a polar transformer network is proposed which combines STN and canonical coordinate representations. In \cite{kim2020cycnn}, a CNN model was proposed which uses polar mapping as a translation module.

Cohen et al. \cite{cohen2016group, cohen2016steerable} collected the benefit of using a subordinate symmetry group that consists of $2\pi$ degree rotations and flipping, and then applied pooling in the group for object detection. This strategy was only used in the deeper layers, as pooling in the shallow layers may result in discarding significant information and declining the system performance. In \cite{ngiam2010tiled}, rather than defining a symmetry group, a pooling method across various untied filters is proposed, which enables the network to estimate the type of invariance. In \cite{cohen2019gauge}, the performance of equivariant networks is extended from global symmetries to local gauge transformation by adopting manifold data learning into the network. However, due to the wide range of parameters, the model is not efficient.  In \cite{cohen2018spherical}, the authors proposed an efficient rotation-equivariant convolution blocks by using a Fast Fourier Transform algorithm. In \cite{marcos2017rotation}, the rotation invariance was integrated into the standard CNN architecture to perform image classification and segmentation. 
In our proposed REFIPN, different from the previous methods, we performed pooling (without adding additional parameters) at various orientations where this process finds a compromise between the model efficiency and the amount of orientation information preserved over the layers. This amendment in the convolution structure provides the possibility to create rotation equivariant models able to detect the influential orientations. On the other hand, by eliminating information of non-maximum orientations, we reduce the scale of feature maps and number of filters to decrease computation costs.

\section{PROPOSED METHOD}
\label{sec:3}
In this section, we first discuss the details of our proposed rotation equivariance model and then detail the Feature Image Pyramid Network. Fig. \ref{fig:2} shows the basic architecture of REFIPN.

\subsection{Vector Field Convolution  for Equivariant Rotation }
Vector field convolution uses an additional factor to generate active models. This additional factor is computed by having a vector field in the edge map is taken from the input image. Vector field convolution uses the standard energy minimization framework, constructed based on an offset condition. The vector field convolution is not only able to capture large concavities, but also has higher robustness to initialization, and has a lower computation cost compared to the standard convolution \cite{li2007active}.

To attain rotation equivariance, we execute the convolution operation with multiple rotated instances of a canonical filter (details are shown in Fig. \ref{fig:3}). The canonical filter $f$ is rotated with respect to various equal space orientations.
To address the problems of invariance or equivariance, we utilize an incremental value $\alpha=[0^{\circ}, 2\pi^{\circ}]$. Nonetheless, this incremental value can be adjusted to a more functional type of deformations. Here, at each distinct point the $f$'s output is computed by using the magnitude of the highest scoring activation achieved via a different range of orientations at the associated angles. By transforming this polar representation towards a Cartesian vector, every single filter $f$ generates a feature map of vector field $V\in \mathbb{R}^{H\times W\times2}$ ($H$ and $W$ represent height and width respectively), in which the output of each point contains two values $[p, q]\in \mathbb{R}^2$ that encodes the highest activation in terms of direction and magnitude. Considering that the feature maps are converted to vector fields, from now on, vector fields are used instead of the convolution filters (as shown in Fig. \ref{fig:3}). Moreover, using Cartesian vectors, the length and breadth constituents $[p, q]$ are orthogonal, which means a vector field convolution is calculated on each constituent by use of standard convolutions. REFIPN contains special key components to process vector fields as inputs and outputs. The subsequent sections illustrate the reformulation of standard convolution layers to be used for filters and vector field activations.\\

\subsubsection{Convolution with Encoding Rotation}
For an image $x\in \mathbb{R}^{H\times W\times c}$  with zero-padding, filter $f\in  \mathbb{R}^{m\times m\times c}$ is implemented at $\Lambda$ orientations with the angles of $\alpha_r=\frac{2\pi}{\Lambda}$, where, $r=1, 2, ..., \Lambda$. Individually, each rotated variation of the canonical filters (blue squares shown in Fig. \ref{fig:3}) is estimated by using bilinear resampling after a particular rotation $\alpha_r$ around the directions of the filter's center as, $f^r=d_{{\alpha_r}}(f)$, in which $d_{\alpha}$ is the rotation operator based on the rotation degrees $\alpha$. In this process, resampling is generally needed excluding the multiplications of $90^{\circ}$ rotations. We have observed that rotations can adjust the weights, and push them towards the edges of the filters even if it is beyond spatial support. The proposed rotation convolution has efficient performance since only the weights inside a circle with diameter $m$ are considered. Therefore, to compute the tensor's output $y\in \mathbb{R}^{H\times W\times \Lambda}$ with $\Lambda$ feature maps, we use, $y^{(r)}=(x\ast f^r)$, in which ($*$) denotes the convolution's operator. $y$ which is a tensor encodes the rotation output space in such a manner that if the input is rotated, the model can cause changes via a series of translation on the generated feature maps. Since, the network only preserves the filters $f$, during the back-propagation operation, each rotated filter's gradients ($\nabla f^r$) are adjusted using the Jordan canonical form as follows:
\begin{eqnarray}
\nabla f= \sum_r d-\alpha_r(\nabla f^r). 
\label{eq:4}
\end{eqnarray}

The proposed model has the capacity to be used on both standard CNN feature maps and vector field feature maps (see Fig. \ref{fig:3}). In comparison with a standard CNN which individually learns $\Lambda$ filters in different orientations, our proposed model generates $\Lambda$ times less parameters to have equal representation. The proposed rotation-convolution model is calculated on all the components independently that leads to 3D tensors defined as, $(V * f)=(V_p * f_p)+(V_q * f_q)$, in which $p$ and $q$ represent the horizontal and vertical components respectively. We should mention that, when $f$ is a 2D vector field, the image rotation operator $d_{\alpha}$ needs to perform another process as follows:

\begin{equation}
\begin{aligned}
f_p^r = cos(\alpha_r)\;d_{\alpha_r}(f_p) - sin(\alpha_r)\;d_{\alpha_r}(f_q)\\
f_q^r = cos(\alpha_r)\;d_{\alpha_r}(f_q) - sin(\alpha_r)\;d_{\alpha_r}(f_p)
\label{eq:6}
\end{aligned}
\end{equation}

\subsubsection{Rotation Equivariant Filter Basis}
As previously discussed, for each point $s\in \mathbb{R}^{H\times W\times 2}$ we assume $K_n$-dimensional feature vectors $V(s)=\oplus_i V^i (s)$ are conducted on the basis of uniform features $V^i (s)$ of dimension $2 l_{in}+1$. In a simple representation, once the object is rotated, two issues occur: the vector from $s$ is transferred to a new (rotated) location $r{+1}$s, and all the vectors are rotated by a $3\times 3$ rotation matrix $M(r)$. Therefore, $M_n (r)$ that works with filters in layer $n$ is block-diagonal, represented as $B^{l_{in}} (r)$ of the $i^{th}$ block. This means the filter $f\in \mathbb{R}^3 \rightarrow \mathbb{R}^{K_{n+1}\times K_n}$ can be divided into blocks $f^{j l}\in \mathbb{R}^3 \rightarrow \mathbb{R}^{(2j+1)\times (2 l+1)}$ for better mapping among features. Each block is bounded to the transformation (rotation) as follows:
\begin{eqnarray}
f^{j l}(rs) = B^j(r) f^{j l}(s) B^l(r)^{-1}
\label{eq:8}
\end{eqnarray}

To have a more compliant form, these filter blocks are vectorized to $V(f^{j l} (s))$, and therefore Eq. (\ref{eq:8}) can be rewritten in the form of a matrix-vector as follows:
\begin{eqnarray}
V(f^{j l}(rs)) = [B^j \otimes B^l](r) V(f^{j l}(rs)), 
\label{eq:9}
\end{eqnarray}
in which we adopt the orthogonality of $B^l$ and the tensor outcome consists of irreducible polynomials. For irreducible polynomials, $B^j\otimes B^l$ can be subdivided in $2min(j, l)+1$ irreducible polynomial order. By determining the changes of basis matrix $Q=(2l+1)(2j+1)\times (2l+1)(2j+1)$, the representation turns into:
\begin{eqnarray}
[B^j \otimes B^l](r) = Q^r[\oplus_{J=\vert j-1\vert}^{j+1}]Q,  
\label{eq:10}
\end{eqnarray}
Consequently, to enforce the constraint, the basis can be amended to $\eta^{jl}(s):= QV(f^{jl} (s))$. Thus Eq. (\ref{eq:8}) becomes,
\begin{eqnarray}
\eta^{j l}(rs)= [\oplus_{J=\vert j-1\vert}^{j+1} B_{(r)}^J]\eta^{j l}(s). 
\label{eq:11}
\end{eqnarray}

Therefore, by using this basis, the block-diagonal configuration of the representation $\eta^{jl}$ is decomposed into $2min(j, l)+1$ subspaces of $2J+1$ dimensions with different constraints:
\begin{eqnarray}
\eta^{j l}(s)= \oplus_{J=\vert j-1\vert}^{j+1}  \eta^{j^{l, J}}(s), \eta^{j^{l, J}} (rs)= B^J(r) \eta^{j^{j, l}}(s). 
\label{eq:12}
\end{eqnarray}

To gain a comprehensive basis, we select different forms of radial basis functions $\varphi^m: \mathbb{R}_+\rightarrow \mathbb{R}$, and filter basis functions are defined as: $\eta^{(j l, J m)} (s)= \varphi^m (\Vert s\Vert) Y^J (\frac{s}{\Vert s\Vert})$, in which $Y$ represents the {\it spherical coordinates} \cite{suda2002fast}. In our experiments, similar to \cite{weiler2018learning}, we also select a Gaussian radial $\varphi^m (\Vert s\Vert)=exp(-\frac{1}{2}[(\Vert s\Vert-m)^2/\sigma^2])$ where $\sigma$ denotes the sigmoid function and the fixed radius of the basis ($j = l= 1$). By associating all $\eta^{(j l, J m)}$ to their primary basis through unvectorization, and $Q^r$, we gain a basis $f^{(j l, J m)}$ for the equivariant filters between the feature spaces (fields) of $j$ and $l$, where the basis is indexed by the frequency $J$ and radial $m$. In the network's forward pass, the basis filters can be linearly integrated as $f^{j l}=\underset{Jm}{\sum}\mathcal{W}^{(j l, J m)} f^{(j l, J m)}$ by adopting learnable weights $(\mathcal{W})$ to build a complete filter $f$, for going into the convolution operations routine.\\

\subsubsection{Max and Orientation Pooling for Vector Fields}
Generally, in CNNs, max-pooling (MP) is used to acquire limited invariance for minor transformations and size reduction of feature maps. Max-pooling is performed by down-sampling of input feature map $x\in \mathbb{R}^{M\times N\times c}$ to $x\in \mathbb{R}^{\frac{M}{w}\times \frac{N}{w}\times c}$, carried out by using the largest value $z$ of each non-overlapping $w\times w$ regions of $x$. It is measured by $x_w [z]= \underset{i\in z}{\max}\; x[i]$, and we can define it more precisely as follows:
\begin{eqnarray}
\mathcal{Y}_w [z]=\mathcal{Y} [j], ~ while ~ j=\underset{i\in z}{argmax}\; \mathcal{Y} [i]
\label{eq:13}
\end{eqnarray}

Consequently, we can define the vector fields' max-pooling as follows:
\begin{eqnarray}
V_w [z]=V [j], ~ while ~ j=\underset{i\in z}{argmax}\; \rho[i]
\label{eq:14}
\end{eqnarray}
in which $\rho$ denotes a scalar map that contains the dimensions of the vectors in $V$.
Furthermore, we apply orientation-pooling (OP) to converting the 3D output tensor $\mathcal{Y}$ to a 2D vector field $V\in \mathbb{R}^{H\times W\times 2}$. This approach prevents the dimensionality issue retaining the activating orientation information of $f$ through a 2D feature map based on the maximal activation ($\rho \in \mathbb{R}^{H\times W}$), and orientations ($\theta \in \mathbb{R}^{H\times W}$). 
\begin{eqnarray}
\rho[i, j] = \underset{r}{argmax}\; \mathcal{Y}[i, j, r],
\label{eq:15}
\end{eqnarray}
\begin{eqnarray}
\theta[i, j] = \frac{2\pi}{\Lambda} \underset{r}{argmax}\; \mathcal{Y} [i, j, r].
\label{eq:16}
\end{eqnarray}

We perform orientation-pooling as a polar coordinate on a 2D vector field which requires $\mathcal{Y}$ before the orientation pooling returns non-negative values $(\rho[i, j] \geq 0)$. All the biases are taken zero as the initial value and ReLU is used as the activation function, determined as $ReLu(x) = argmax(x, 0)$, to $\rho$, which provides stable training. In terms of weight initialization, we propose to normalize their weights variance with an extra factor of $\Lambda$. To satisfy rotation transformation, we transform the canonical coordinate system into the Cartesian form as follows:
\begin{eqnarray}
p=cos(\theta) ReLU (\rho) 
\label{eq:17}
\end{eqnarray}
\begin{eqnarray}
q=sin(\theta) ReLU (\rho)
\label{eq:18}
\end{eqnarray}
where $p, q \in \mathbb{R}^{H\times W}$ and the vector field ($V$ ) is formulized as:
\begin{eqnarray}
\begin{bmatrix}
1\\0
\end{bmatrix}
p+ 
\begin{bmatrix}
0\\1
\end{bmatrix}
q
\label{eq:19}
\end{eqnarray}
\subsubsection{Vector Fields Batch Normalization (VFBN)}
Batch normalization (BN) \cite{ioffe2015batch} normalizes all the feature maps in a mini-batch to a form of zero mean and mean square. It enhances convergence by stochastic gradient descent training.
Along with vector fields and orientation of activations, we use BN to normalize the weights of the vectors to mean square deviation. It is not required to normalize the gradients, because their values are limited and bounding their distribution may result in losing essential information of relative orientations. By using a feature map of a vector field $V$ and its scalar map $\rho$, the VFBN is calculated as follows:
\begin{eqnarray}
VFBN = \frac{V}{\sqrt{var(\rho)}}
\label{eq:20}
\end{eqnarray}
where $var(\rho)$ denotes the variance of the scalar map.

\subsection{Rotation Equivariant Feature Image Pyramid Network}
In this section, we describe the architecture of our oriented object detector by adopting the proposed $rotation-convolution\; (RConv)$.
In our approach, we use the standard SSD \cite{liu2016ssd} as the baseline detector. As earlier stated, the SSD traces objects in a pyramid network by using several CNN layers, in which each layer is appointed for detecting particular scale objects. This means, small objects are detected by shallower layers that have small receptive fields, while deeper layers that have larger receptive fields are used in order to detect larger objects. Nonetheless, the SSD, because of lack of information in the shallow layers, fails to correctly detect tiny objects \cite{huang2017speed}. To address this issue of SSD, we build a feature pyramid network from image pyramids to improve the SSD's detection performance without affecting its speed. 

As shown in Fig. \ref{fig:2}, REFIPN contains two key modules: the conventional SSD network plus the light-weight image pyramid module (LIPM) for generating semantically strong features. Similar to SSD, VGG-16 is adopted as the backbone and we insert several progressively smaller convolution layers to improve feature extraction. Differing from SSD, REFIPN contains a LIPM's layer in different layers of SSD by using an attention module. Then, the features of each layer are fused with the features of the previous layer using a fusion module.\\

\subsubsection{Light-weight Image Pyramid Module (LIPM)}
As previously reported, the conventional feature pyramid networks (FPNs) \cite{lin2017feature, pang2019efficient} are not efficient since different scales of each image go through a CNN to extract feature maps of each scale. To address this issue, we introduce an efficient solution to generate object candidate by use of RPN in LIPM \cite{ren2016faster, ding2019learning}. The network consists of a constant down-sampling process by adopting $RConv\; layers$. The LIPM receives feature maps of an arbitrary size as input and outputs a group of box offsets. On the bases of box offset sizes, the network selects a feature map layer in the most fitting scale. For each input image $X$, first an image pyramid $X_p$ is created using multiple down-sapling processes as $X_p=\{x_1, x_2, ..., x_n\}$, where, $n$ represents the number of image pyramid layers and each particular scale of the LIPM matches a layer of the SSD prediction layer. To generate multi-scale feature maps, all of the image scales go over a LIPM $S_p=\{s_1, s_2, ..., s_n\}$, in which, $S_p$ indicates the features of different layers. The LIPM contains two $3\times 3$ and a $1\times1$ $RConv$ layers with different numbers of channels to match the result of the LIPM with the $RConv$-SSD feature maps.\\

\subsubsection{Spatial Attention Module}
To insert the generated features of the Light-weight RConv layers into the SSD layers, we use a spatial attention module, as illustrated in Fig. \ref{fig:2}(c). In the first step, both the features from the matching layers of the LIPM and the SSD are processed through a VFBN layer and $1\times1$ $RConv$ layer. Next, we use element-wise concatenation to combine the normalized features. Moreover, $3\times 3$ and $1\times1$ $RConv$ layers are adopted to reform features. For an input image $I$, features $\acute{s}_n$ from the prediction layer of SSD $n^{th}$ are merged with the corresponding light-weight features $s_n$ as, $r_n= \delta_n (\beta(\acute{s}_n)\otimes \beta(s_n))$, in which, $r_n$ are the reformed features after merging, $\delta_n (.)$ represents the process of $1\times1$ and $3\times3$ $RConv$ layers, and $\beta(.)$ indicates the VFBN procedure.\\
In our network, the SSD takes the ROIs from RPN as input which helps to improve object localization and detection \cite{ding2019learning, zheng2020hynet}. The LIPM is connected to the convolution layers of SSD via spatial attention module.\\ 

\subsubsection{Feature Fusion Module}
\begin{figure}
  \centering
  \includegraphics[width=0.48\textwidth]{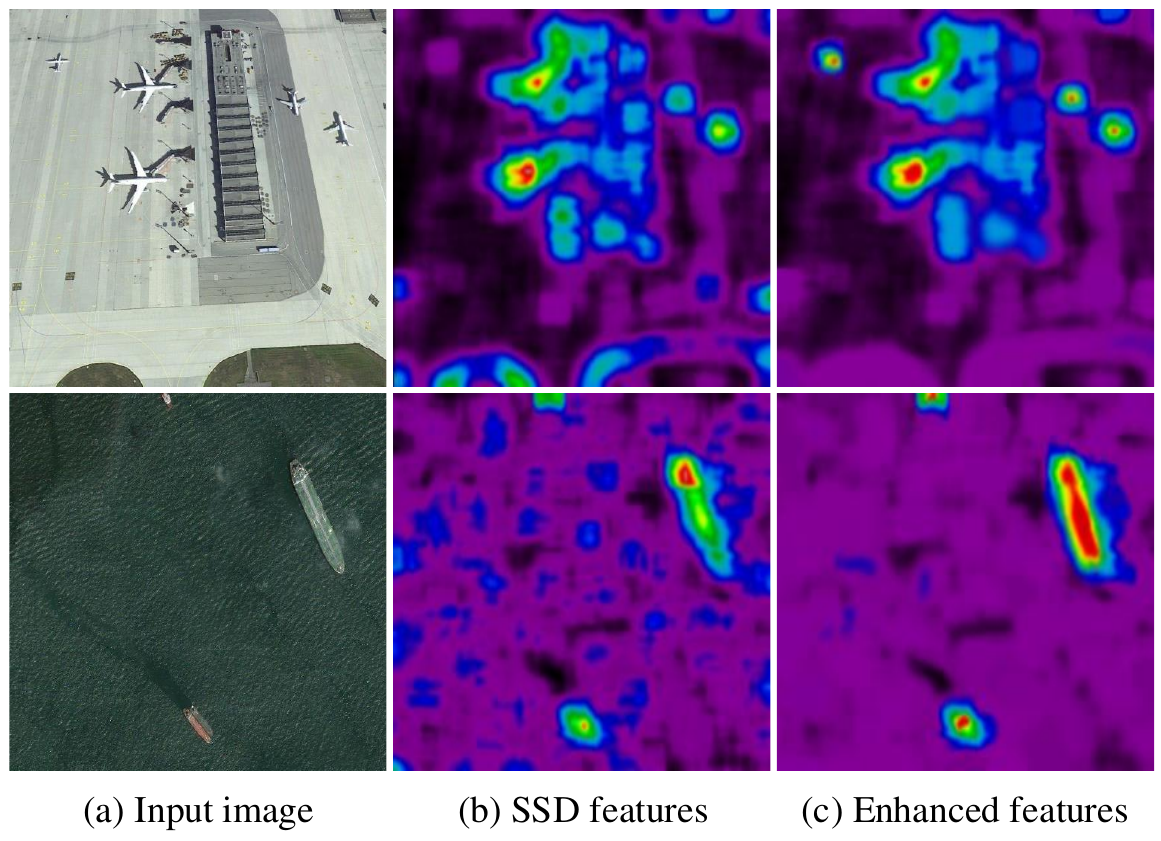}
\caption{\small Comparison of feature maps that acquired from the $(Conv4)$ layer of the conventional SSD and $(RConv4)$ layer of our model after the feature fusion module.}
\label{fig:4}
\end{figure}

In order to further improve the spatial information, we propose a light-weight feature fusion module (FFM) to normalise the weights and combine features from both the current and previous layers (Fig. \ref{fig:2}(d)). Through FFM, first, both the current and former layers are passing through a $1\times1$ $RConv$ layer to produce the same size of data. Then, previous $r_{n-1}$ and current $r_n$ features are integrated through an element-wise addition. This feature enhancement operation is followed by $3\times3$ and $1\times1$ $RConv$ layers to produce the final detection $d_n= \Upsilon (\phi(r_{n-1})\oplus \phi_n (r_n))$, in which, $\phi_n(.)$ represents the operation including the serial $1\times1, 3\times3$ $RConv$ and VFBN layers, and $\Upsilon$ is the ReLU activation operation. Fig. \ref{fig:4} shows the effect of our proposed REFIPN to enhance the discriminative features of SSD.

\section{EXPERIMENTS AND ANALYSIS}
\label{sec:4}
In this section, we describe our experimental details, such as the datasets used in our experiments, followed by evaluation metrics, and we perform various experiments to evaluate the performance and efficiency of our proposed model. In addition, an extensive ablation study is conducted to evaluate the performance of each proposed module. We used a Tesla V100 GPU to implement and evaluate the networks. The model is implemented using Keras $2.1.2$, and TensorFlow $1.3.0$ GPU as the backend deep learning engine. 

\begin{table}
\centering
  \caption{\small{ARCHITECTURE {\footnotesize{OF}} {\footnotesize{THE}} REFIPN WITH $\Lambda = 17$ SAMPLED ORIENTATIONS. THE VGG ARCHITECTURE NOT INCLUDED.}}
  \label{tab:1}
  \begin{tabular}{l|c|c}     \hline 
 Operation & Filter Size & Feature Channels   \\     \hline
 RConv5, OP & $3\times 3$ &  512  \\ 
 MP   & $3\times 3$ & -   \\ \hline 
 RConv6, OP & $3\times 3$ &  1024  \\ \hline
  RConv7, OP   & $1\times 1$ &  1024  \\ \hline
 RConv8, OP & 
 \begin{tabular}{c}   $1\times 1$ \\ $3\times 3$ , stride=2 \\
 \end{tabular} & 
 \begin{tabular}{c}   256 \\ 512 \\
 \end{tabular} \\ \hline
  RConv9, OP & 
 \begin{tabular}{c}   $1\times 1$ \\ $3\times 3$, stride=2 \\
 \end{tabular} & 
 \begin{tabular}{c}   128 \\ 256 \\
 \end{tabular} \\ \hline
 RConv10, OP & 
   \begin{tabular}{c}   $1\times 1$ \\ $3\times 3$ \\
 \end{tabular} & 
 \begin{tabular}{c}   128 \\ 256 \\
 \end{tabular} \\ \hline
 RConv11, OP & 
   \begin{tabular}{c}   $1\times 1$ \\ $3\times 3$\\
 \end{tabular} & 
 \begin{tabular}{c}   128 \\ 256 \\
 \end{tabular} \\ \hline
\end{tabular}
\end{table}
\subsection{Datasets and Evaluation Metrics}
{\it DOTA} \cite{xia2018dota}. It is a large RSI dataset used for object detection which comprises of $2806$ images with different size ranges ($800\times 800$ to $4000\times 4000$) and $15$ classes of objects with different orientations, and scales.

{\it NWPU VHR-10} \cite{dong2019sig}. This is a 10 class and manually annotated dataset that contains $650$ optical RSIs, in which $565$ images were obtained from Google Earth and have various sizes ($533\times 597$ to $1728\times 1028$ pixels). We randomly selected $70\%$ of the original images to form the training set, $10\%$ as the validation set, and the rest as the testing set.

For the DOTA dataset in the training phase, we split the images into the $400 \times 400$ pixels sub-images with 200 pixels overlap between the neighboring sub-images. In the testing phase to evaluate the effect of input size in the detection results, we prepare two sets of the testing dataset. Image patches of $400 \times 400$ and $800 \times 800$ pixels are cropped from the test set images with 200 and 400 pixels overlaps respectively. We also use the multi-scale technique in this process \cite{xu2020hierarchical}. More precisely, first we rescale the original images by $0.8\times$, and $0.4\times$ before splitting, and then take all the patches as the training and testing sets. Since DOTA contains large number of images, we randomly selected 19,853 patches for training, 1836 (400 pixels) and 1570 (800 pixels) patches for testing.

For the NWPU VHR-10 dataset, the quantity of images is insufficient for training, to increase the training set, we perform rotation, rescaling, and mirroring. Furthermore, as the number of the entities in each class was unbalanced, we use different techniques for each class to balance the number of objects. Similar to DOTA for the NWPU VHR-10, we rescale the original images by $0.5\times$ and $1.5\times$ before splitting, and we prepared two test set images ($400 \times 400$ and $800 \times 800$ pixels) for evaluating the detection performance of state-of-the-art approaches over different input sizes. 

In addition, to evaluate the performance of the state-of-the-art models in the estimation of objects orientation, we prepare another dataset by using the plane and car categories of the original DOTA dataset. We crop a $80 \times 80$ square patch around each plane and car, according to central location of the bounding box. These cropped patches are used for training and testing of the baseline models. 834 planes and 729 cars are used for training and 160 planes and 130 cars for testing. 
\begin{table}[b]
\centering
  \caption{AVERAGE ERROR IN THE ESTIMATION OF AIRPLANE AND CAR ORIENTATIONS. }
  \label{tab:2}
  \begin{tabular}{l|c|c|c}     \hline
Methods & Airplane error $^{\circ}$ & Car error $^{\circ}$ & param  \\     \hline
CNN \cite{jaderberg2015spatial}	& 25.71 & 29.83 & 27k  \\
STN \cite{jaderberg2015spatial} & 20.34 & 25.67 & 21k  \\
GEN \cite{cohen2016group} & 17.84 & 23.96	& 14k   \\
ORSIm \cite{wu2019orsim} & 15.69 & 22.35 & 23K   \\
NSS \cite{jiang2020rotation} & 15.31 & 21.85	& 15k   \\
LR-CNN \cite{guo2020rotational}	& 14.52 & 21.17 & 26k   \\
REFIPN & 13.37 & 19.42 & 9k   \\
        \hline
\end{tabular}
\end{table}

For evaluating the performance of our proposed detector, the frames per second (FPS) and mean average precision (mAP) are used as evaluation metrics. mAP is computed as follows:
\begin{eqnarray}
mAP=\int_{0}^{1} P(R)\; dR,
\label{eq:23}
\end{eqnarray}
in which $P$ and $R$ denote the estimated precision and recall rates respectively, and $d$ represents the value of the estimated center point's coordinates. For fair comparison with other approaches, we only use random horizontal flipping in the training to avoid over-fitting.\\
In \cite{fu2020point, fu2020rotation}, metrics are introduced to evaluate the localization and classification ability of methods. \\
{\it Localization Error} is determined as the gap between the corners of the detection box and the ground-truth, and the value is normed by the size of object. The metrics contain means of localization error on top of horizontal and vertical axis and standard deviations of localization error beside influenced axes. In this metric, if a false positive (FP) sample is accurately detected but has slight overlap ($0.1 < IoU < 0.5$) with ground truth, it is categorized as localization error. Here, the $IoU$ signifies the intersection area of two oriented boxes.\\
{\it Confusions With Background (BG)} is determined if an FP sample has unsatisfactory overlap ($IOU < 0.1$) with any object, it is classified as a confusion with BG.
\begin{figure}
\centering
  \includegraphics[width=0.48\textwidth]{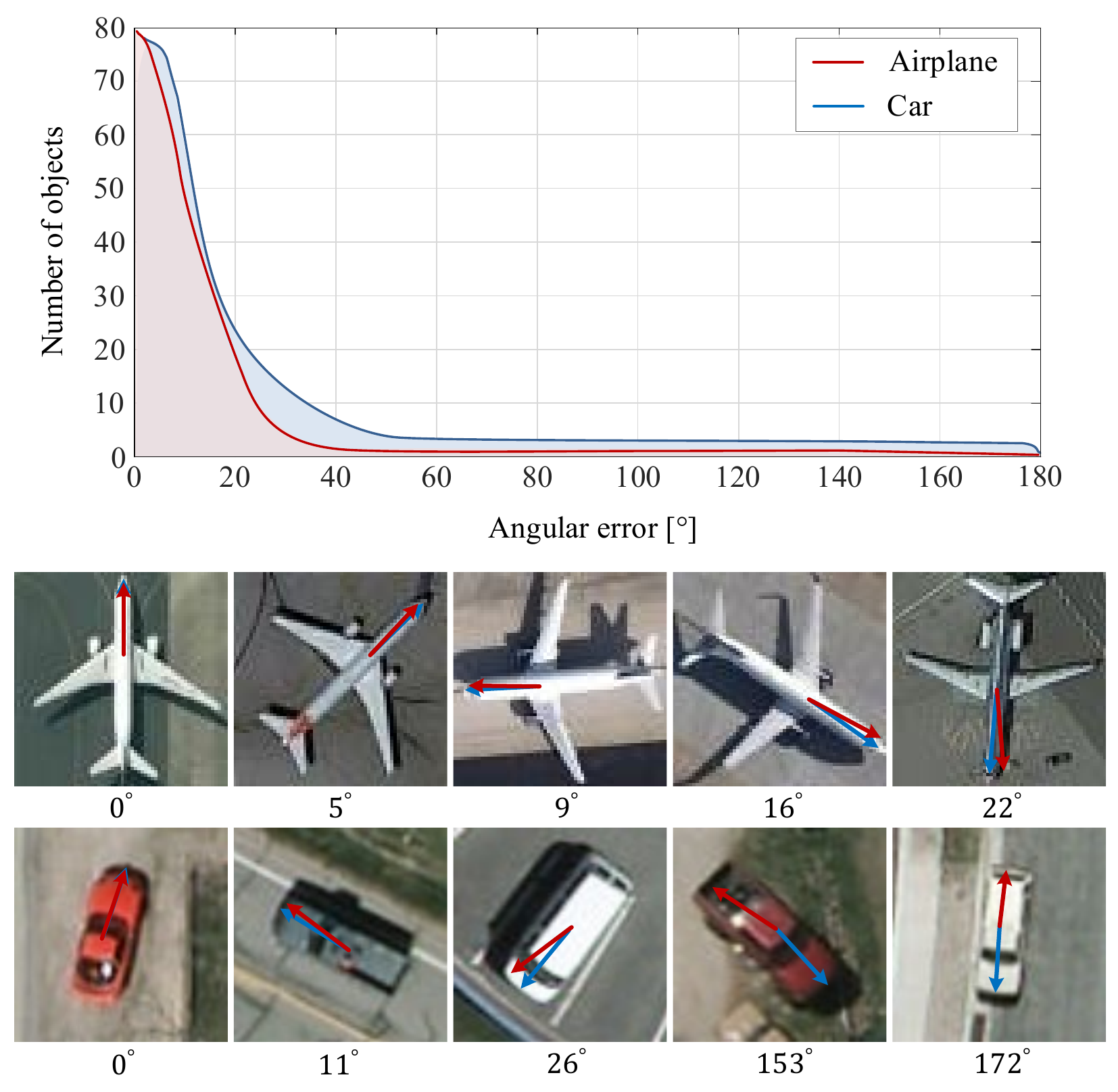}
\caption{\small The errors distribution in the test set (top). Bottom shows the samples of accurately and inaccurately predicted orientations. Ground truth arrows in \textcolor{blue}{blue} and estimations in \textcolor{red}{red}.}
\label{fig:5}
\end{figure}
\subsection{Implementation Details}
We wish to learn an equivariant rotation function, which means $\Delta\alpha^{\circ}$ rotation in the input image leads to $\Delta\alpha^{\circ}$ transform in the estimated gradient. In general, we train on $sin$ and $cos$ of $\alpha^{\circ}$, as they both are constants in terms of $\Delta\alpha^{\circ}$. Table \ref{tab:1} shows the network's architecture of the proposed detector. Based on SSD \cite{liu2016ssd}, VGG-16 is adopted as the backbone network pretrained on the ILSVRC CLS-LOC dataset. The SSD utilizes $(RConv4)$ and fully connected ($FC7$, transformed to a $RConv$) layers from the standard VGG-16 network. It reduces the endmost FC layer of the VGG-16 architecture and inserts various ranges of smaller conv layers: $[RConv8,..., Rconv11]$, with different feature sizes and features of different layers' LIPM are integrated with their corresponding layers in the SSD network ($s_1$ for $(RConv4)$ and $s_n$ for the last layer $(RConv9)$).
\begin{figure*}
  \centering
  \includegraphics[width=0.95\textwidth]{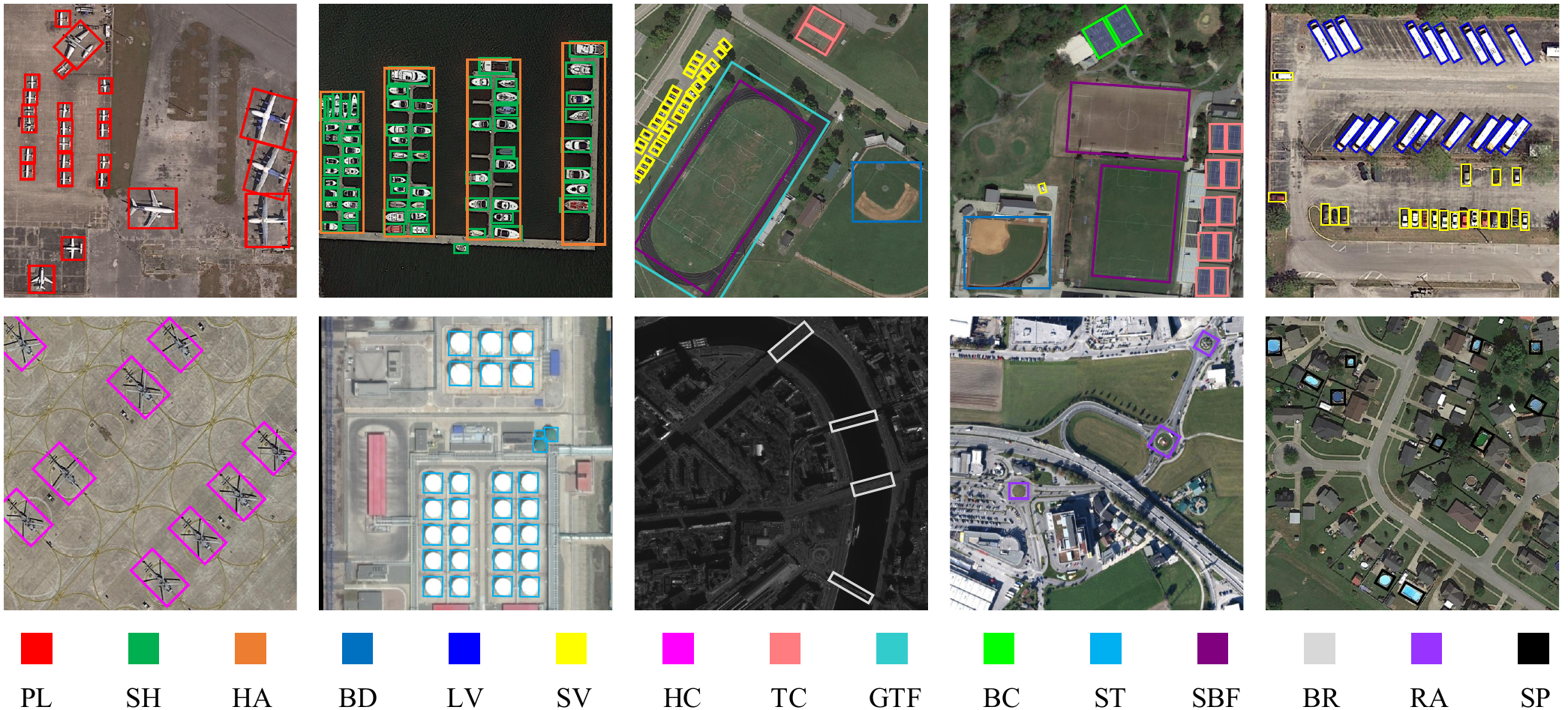}
\caption{Exemplar detection results on the test DOTA dataset. Plane (PL), Ship (SH), Harbor (HA), baseball diamond (BD),  Large vehicle (LV), Small vehicle (SV), Helicopter (HC), Tennis court (TC), Ground track field, (GTF), Basketball court (BC), Storage tank (ST), Soccer ball field (SBF), Bridge (BR), Roundabout (RA), and Swimming pool (SP). }
\label{fig:7}
\end{figure*}
\begin{figure}
\centering
  \includegraphics[width=0.44\textwidth]{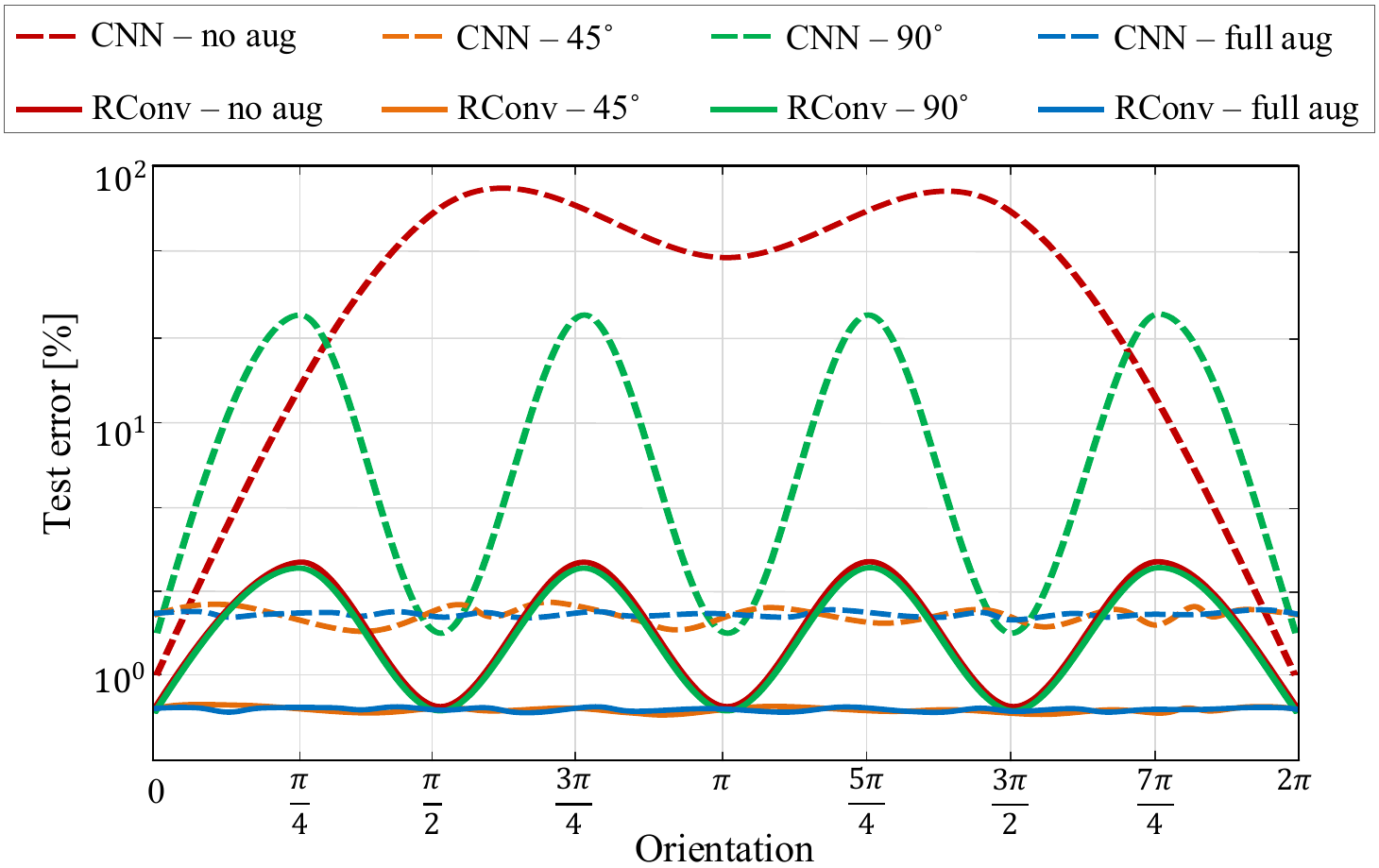}
\caption{Comparison of rotational generalization abilities of a standard CNN and a RConv with $\Lambda = 17$. In this evaluation the train set is composed of un-rotated objects while for the test set the objects are rotated to an identical angle.}
\label{fig:6}
\end{figure}

For the output, the output vector normalization is used along with a non-linear tanh activation. In the first layer of SSD with $RConv$, we learn $C = 3$ filters, in $\Lambda=17$ orientations, which refers to $C \Lambda=51$ active channels. We expect such vectors to go through a circular transformation once the input image undergoes a rotation. The mappings of the last layer is as follows $[sin(2\pi/\Lambda), sin(2\times 2\pi/\Lambda), ..., sin(\Lambda \times 2\pi/\Lambda)]$ and $[cos(2\pi/\Lambda), cos(2\times 2\pi/\Lambda), ..., cos(\Lambda \times 2\pi/\Lambda)]$. This process guarantees that all the preferred orientations are detectable by our proposed model. Fig. \ref{fig:5} demonstrates the error distribution of the test set. Majority of the samples, 91.4\% of airplanes and 81.6\% of cars, are estimated with less than $14^{\circ}$ of the orientation error, where the large amount of the total error rates 6.9\% are due to the samples whose error rates are above $160^{\circ}$, which means, the front side of the object has been incorrectly taken instead of its rear side.\\ 
NWPU VHR10 labels are in a conventional axis-aligned bounding boxes (BBs) form, while DOTA objects' labels are in a quadrilateral form. For adopting the both settings, our proposed REFIPN uses both horizontal and oriented BBs (HBB, OBB) as ground truth, where HBB:$\{x_{min},y_{min},x_{max},y_{max}\}$, OBB:$\{x_{center},y_{center},w,h,\theta \}$, here $w$, $h$ denote width, height and $\theta$ is within $[0, 90^{\circ})$ for each object. In training, the OBB ground truth is produced by a group of rotated rectangles which properly overlap with the given quadrilateral labels. For the NWPU-VHR-10 datasets, REFIPN just produces HBB detection results, because OBB ground truth does not exist in the datasets. However, for the DOTA, REFIPN produces both OBB and HBB outputs, as presented in Fig. \ref{fig:7}.
In our model, for the loss function, we follow Faster RCNN \cite{ren2016faster}. In the proposed REFIPN, the loss function is constructed as $\ell_{REFIPN}=\ell_{rpn}+\ell_{head}$, in which $\ell_{rpn}$ represents the loss of RPN introduced in \cite{ren2016faster} and $\ell_{head}$ indicates the heads' loss for Faster RCNN formulated as follows:
\begin{equation}
 \begin{aligned}
\ell_{rpn}&=\lambda_1 \frac{1}{N_{cls}}\sum_i \ell_{cls}(p_i, p^{\ast}_i) \\ 
&  +\lambda_2 \frac{1}{N_{reg}}\sum_i p^{\ast}_i \ell_{reg}(t_i, t^{\ast}_i)
 \end{aligned}
\end{equation}

\begin{equation}
 \begin{aligned}
~~~~~~\ell_{head}&=\lambda_3 \frac{1}{N^{\ast}_{cls}}\sum_i \ell_{cls}(c_i, c^{\ast}_i) \\
&  +\lambda_4 \frac{1}{N^{\ast}_{reg}}\sum_i [c^{\ast}_i \geq 1]\ell_{reg}(h_i, h^{\ast}_i) \\
&  + \lambda_5 \frac{1}{N^{\ast}_{reg}}\sum_i [c^{\ast}_i \geq 1]\ell_{reg}(o_i, o^{\ast}_i)
 \end{aligned}
\end{equation}
\begin{table*}
\centering
  \caption{DETECTION ACCURACY AND SPEED COMPARISONS OF OUR MODEL WITH EXISTING RSI DETECTORS ON DOTA TEST SET FOR HBB and OBB TASK. WITH EXACTLY THE SAME SETTINGS EXCEPT FOR DIFFERENT INPUT SIZES ($400 \times 400$ AND $800 \times 800$) WE EVALUATE EACH MODEL PERFORMANCES. THE BEST RESULTS of 400 PIXEL INPUT SIZE ARE HIGHLIGHTED IN {\textcolor{red}{RED}} AND 800 PIXEL IN {\textcolor{blue}{BLUE}}. OUR DETECTOR SHOWS AN IDEAL TRADE-OFF BETWEEN SPEED AND DETECTION IN COMPARISON WITH THE OTHER DETECTORS}
  \label{tab:3}
  \begin{tabular}{m{5.2em} m{1.8em}|  m{0.5cm} m{0.5cm} m{0.5cm} m{0.5cm} m{0.5cm} m{0.5cm} m{0.5cm} m{0.5cm} m{0.5cm} m{0.5cm} m{0.5cm} m{0.5cm} m{0.6cm} m{0.5cm} m{0.5cm}| m{0.5cm} m{0.27cm}}     \hline
    Methods &  Input&Plane&BD&Bridge&GFT&SV&LV&Ship&TC&BC&ST&SBF&RA&Harbor&SP&HC&mAP&FPS  \\     \hline
\multicolumn{19}{c}{\it Detectors for HBB task}  \\ 
\hline
\multirow{2}{*}{SSD \cite{liu2016ssd}} &	$400$ &85.61&79.48&47.59&65.03&66.54&71.62&73.41&86.62&79.25&70.59&46.57&62.43&66.87&56.38&57.93& 	68.54	& {\textcolor{red}{134}}  \\
 &	                     $800$ &86.67&80.32&48.11&65.35&67.18&72.33&74.45&87.41&80.64&71.22&47.36&63.26&67.49&57.36&58.89& 	69.23	& {\textcolor{blue}{68}} \\ \hline
\multirow{2}{*}{SSD + FPN} & 	$400$ &86.43&79.89&47.35&65.80&67.28&72.39&73.62&86.35&79.96&70.86&47.52&62.84&67.71&56.49&57.38& 68.12		& 87 \\ 
         & 	$800$ &87.95&81.63&50.17&66.73&69.08&73.62&75.44&87.83&81.07&72.38&48.64&64.23&68.96&58.09&59.17& 	69.67	& 44 \\ \hline
\multirow{2}{*}{RICA \cite{li2017rotation}} &	$400$ &84.39&78.36&45.29&66.32&64.71&69.23&73.94&85.12&78.86&70.18&50.24&62.57&69.08&75.51&54.26& 68.59	& 63 \\
&	                           $800$ &86.97&80.93&46.68&67.47&66.19&71.56&74.33&86.43&80.37&71.42&51.76&64.78&71.35&76.84&56.11& 70.21	& 31 \\\hline
\multirow{2}{*}{ORSIm \cite{wu2019orsim}} &	$400$ &85.72&79.85&45.20&67.04&66.83&72.35&73.57&85.92&79.42&71.12&51.68&64.89&72.02&76.55&58.09&	70.03	& 79 \\
 &	                     $800$ &87.25&81.12&47.84&68.91&69.27&73.38&75.86&87.39&81.65&72.53&53.08&66.22&73.47&78.94&59.23&	71.74	& 40 \\ \hline
\multirow{2}{*}{FMSSD \cite{wang2019fmssd}} &	
        $400$ &87.75&81.30&48.17&69.88&67.11&72.41&75.53&{\textcolor{red}{89.62}}&82.58&74.28&53.59&65.41&73.28&77.49&58.03&71.76	&44 \\
       &$800$ &89.20&83.58&49.25&70.04&69.34&74.75&77.95&{\textcolor{blue}{90.78}}&83.76&75.39&55.44&67.59&75.39&{\textcolor{blue}{80.85}}&60.41&	73.58&21\\ \hline
\multirow{2}{*}{LR-CNN \cite{guo2020rotational}} &	
     $400$ &86.69&83.27&56.42&75.89&74.88&77.51&83.17&87.26&84.03&82.41&65.12&67.96&81.84&77.38&64.81&	76.59&65 \\
 &	$800$ &88.23&84.91&58.65&77.53&76.63&79.34&{\textcolor{blue}{85.63}}&89.74&85.55&83.15&67.75&69.39&82.29&79.82&66.37&78.33&32 \\ \hline
\multirow{2}{*}{HSP \cite{xu2020hierarchical}} &	$400$            &87.17&84.21&{\textcolor{red}{60.15}}&77.96&76.03&79.35&81.46&87.95&85.79&{\textcolor{red}{85.28}}&{\textcolor{red}{68.31}}&70.61&81.76&79.24&{\textcolor{red}{68.59}}& 78.25 &37 \\
 &	$800$ &89.25&86.40&{\textcolor{blue}{62.54}}&79.37&77.41&81.12&83.92&90.42&87.06&85.54&{\textcolor{blue}{70.51}}&71.88&83.90&80.69&{\textcolor{blue}{69.94}}& 80.22 &19 \\ \hline
\multirow{2}{*}{REFIPN} &	$400$ &{\textcolor{red}{89.91}}&{\textcolor{red}{86.96}}&60.13&{\textcolor{red}{79.63}}&{\textcolor{red}{77.69}}&{\textcolor{red}{80.69}}&{\textcolor{red}{83.71}}&89.53&{\textcolor{red}{86.65}}&85.21&68.26&{\textcolor{red}{71.36}}&{\textcolor{red}{82.59}}&{\textcolor{red}{79.82}}&68.56& 	{\textcolor{red}{79.54}}	& {112} \\
         &	$800$ &{\textcolor{blue}{90.41}}&{\textcolor{blue}{87.63}}&61.91&{\textcolor{blue}{80.50}}&{\textcolor{blue}{78.46}}&{\textcolor{blue}{81.93}}&84.35&90.34&{\textcolor{blue}{87.79}}&{\textcolor{blue}{86.28}}&69.65&{\textcolor{blue}{72.54}}&{\textcolor{blue}{84.98}}&80.56&69.08& 	{\textcolor{blue}{80.43}}	& {57} \\
        \hline
\multicolumn{19}{c}{{\it Detectors for OBB task}}  \\
        \hline
\multirow{2}{*}{ORSIm \cite{wu2019orsim}} &	$400$ &83.61&78.13&39.21&60.36&59.56&63.02&71.84&84.07&78.11&70.36&50.18&62.96&66.84&68.24&56.81&66.22		& 79 \\
 &	                     $800$ &85.39&80.74&40.56&61.75&62.46&64.14&73.69&86.54&80.48&71.12&51.16&64.37&68.69&69.61&58.45&67.94		& 40 \\ \hline
 \multirow{2}{*}{RoITransf. \cite{ding2019learning}} &	$400$ &87.19&76.73&41.47&74.32&67.12&72.14&{\textcolor{red}{82.27}}&88.93&76.74&80.79&57.14&51.44&60.02&58.11&47.18& 68.10	& 71 \\
&	                           $800$ &88.33&78.40&43.19&75.79&68.58&72.56&{\textcolor{blue}{83.54}}&90.76&77.30&81.39&58.35&53.49&62.70&58.67&47.61& 69.37	& 35 \\\hline
\multirow{2}{*}{LR-CNN \cite{guo2020rotational}} &	
     $400$ &86.15&82.63&54.85&70.51&68.91&{\textcolor{red}{72.69}}&81.34&86.74&82.87&81.63&64.47&66.81&75.92&71.02&64.09&	74.04&65 \\
 &	$800$ &87.96&84.63&56.74&71.65&71.31&{\textcolor{blue}{74.96}}&83.41&88.92&85.16&82.73&66.93&68.84&76.37&72.81&65.49&75.86&32 \\ \hline
\multirow{2}{*}{HSP \cite{xu2020hierarchical}} &	$400$            &86.83&84.05&{\textcolor{red}{57.34}}&74.65&71.92&70.58&80.04&87.86&85.67&{\textcolor{red}{85.24}}&{\textcolor{red}{68.07}}&70.22&76.42&72.93&{\textcolor{red}{67.27}}& 75.94 &37 \\
 &	$800$ &89.03&86.36&{\textcolor{blue}{58.38}}&75.86&73.16&73.38&82.31&90.08&87.01&{\textcolor{blue}{85.49}}&{\textcolor{blue}{69.76}}&71.79&78.61&74.84&{\textcolor{blue}{68.83}}& 77.68 &19 \\ \hline
\multirow{2}{*}{REFIPN} &	$400$ &{\textcolor{red}{89.11}}&{\textcolor{red}{84.85}}&56.81&{\textcolor{red}{75.79}}&{\textcolor{red}{73.05}}&72.51&81.92&{\textcolor{red}{89.34}}&{\textcolor{red}{86.27}}&84.18&68.05&{\textcolor{red}{70.35}}&{\textcolor{red}{76.48}}&{\textcolor{red}{74.15}}&67.12& 	{\textcolor{red}{76.66}}	& {112} \\
         &	$800$ &{\textcolor{blue}{90.25}}&{\textcolor{blue}{86.98}}&57.93&{\textcolor{blue}{77.18}}&{\textcolor{blue}{74.32}}&74.60&83.42&{\textcolor{blue}{90.91}}&{\textcolor{blue}{87.25}}&85.40&69.18&{\textcolor{blue}{71.96}}&{\textcolor{blue}{78.95}}&{\textcolor{blue}{75.88}}&68.71& 	{\textcolor{blue}{78.19}}	& {57} \\
        \hline
\end{tabular}
\end{table*}
\begin{table*}
\centering
  \caption{PERFORMANCE COMPARISONS BETWEEN THE BASELINES AND OUR DETECTOR FOR MULTI-SCALE OBJECT DETECTION ON NWPU VHR-10 TEST DATA SET AND OVERALL MEAN AP.}
  \label{tab:5}
   \begin{tabular}{m{5em} m{2em}| p{0.6cm} p{0.6cm} p{0.6cm} p{0.6cm} p{0.6cm} p{0.6cm} p{0.6cm} p{0.7cm} p{0.7cm} p{0.7cm}| p{0.6cm} p{0.5cm}}     \hline
    Methods & Input &Plane&SH&ST&BD&TC&BC&GTF&Harbor&Bridge&Vehicle& mAP & FPS  \\     \hline
 
\multirow{2}{*}{SSD \cite{liu2016ssd}} &	$400$ &87.42&77.93&79.73&83.41&71.89&70.46&85.72&67.49&65.38&62.53& 75.15 & {\textcolor{red}{141}}  \\
                     &	$800$ &89.16&79.51&81.25&85.63&72.55&73.12&87.46&68.21&67.19&65.14& 77.32 & {\textcolor{blue}{71}}  \\ \hline
\multirow{2}{*}{SSD + FPN} & 	$400$ &90.03&81.25&82.63&86.30&76.52&79.59&85.16&70.49&67.35&69.36& 78.48 & 93 \\ 
             & 	$800$ &92.82&83.41&84.46&88.52&78.41&81.77&88.43&72.12&69.28&71.47& 80.76 & 46\\ \hline

\multirow{2}{*}{RICA \cite{li2017rotation}} &	$400$ &95.42&84.25&88.54&90.63&84.14&77.35&88.06&74.18&68.59&73.03& 82.41 & 69 \\
                          &	$800$ &96.36&85.71&89.38&91.47&85.66&78.28&89.27&75.43&69.88&74.92& 83.64 & 35 \\   \hline
\multirow{2}{*}{ORSIm \cite{wu2019orsim}} &	$400$ &97.15&85.02&85.17&91.20&79.57&89.64&92.76&68.50&73.62&72.24& 83.75 & 86 \\
                     &	$800$ &98.21&88.15&88.41&94.33&82.78&92.73&95.07&71.63&75.26&77.69& 86.42 & 43 \\ \hline

\multirow{2}{*}{FMSSD \cite{wang2019fmssd}} &	$400$ &98.53&87.20&87.76&95.47&83.02&{\textcolor{red}{94.35}}&97.22&72.54&78.23&85.59& 87.82 & 53 \\
                          &	$800$ &99.62&88.61&89.52&97.11&84.57&{\textcolor{blue}{95.24}}&98.53&73.68&79.32&87.47& 89.35 & 26 \\ \hline

\multirow{2}{*}{LR-CNN \cite{guo2020rotational}} &	$400$ &97.13&{\textcolor{red}{90.57}}&94.71&95.39&85.23&90.15&94.69&80.03&77.21&84.19& 88.97 & 74\\
                                &	$800$ &98.39&{\textcolor{blue}{91.20}}&95.46&96.23&86.34&91.28&96.64&81.75&78.83&85.94& 90.31 & 38 \\ \hline

\multirow{2}{*}{HSP \cite{xu2020hierarchical}} &	$400$ &98.26&89.24&94.09&{\textcolor{red}{96.93}}&87.20&88.01&97.26&84.74&{\textcolor{red}{85.04}}&85.42& 90.57 & 45 \\
                          &	$800$ &99.45&91.15&96.08&{\textcolor{blue}{97.81}}&88.79&90.34&98.41&{\textcolor{blue}{86.50}}&{\textcolor{blue}{86.92}}&87.85& 92.35 & 23 \\ \hline

\multirow{2}{*}{REFIPN} &	$400$ &{\textcolor{red}{98.85}}&90.39&{\textcolor{red}{96.31}}&96.53&{\textcolor{red}{88.09}}&93.63&{\textcolor{red}{98.36}}&{\textcolor{red}{85.32}}&84.91&{\textcolor{red}{87.14}}& {\textcolor{red}{91.94}} & 118 \\
     &	$800$ &{\textcolor{blue}{99.52}}&91.07&{\textcolor{blue}{97.12}}&97.36&{\textcolor{blue}{89.45}}&94.71&{\textcolor{blue}{98.95}}&86.05&86.40&{\textcolor{blue}{88.21}}& {\textcolor{blue}{92.86}} & 59\\ \hline

\end{tabular}
\end{table*}
in which $i$ denotes the index of a BB, $\ell_{cls}$ represents the cross-entropy loss, $p_i$ is the probability of the $i^{th}$ anchor predicted as an object, $p^{\ast}_i$ denotes the ground-truth category assigned to the $i^{th}$ anchor (1 if the box is positive and 0 for negative one), $\ell_{reg}$ represents the smooth-$\ell_1$ loss, $t_i$ denotes the detected regression offset for $i^{th}$ anchor and $t^{\ast}_i$ represents the target BB regression offset for the $i^{th}$ positive anchor. $c_i$ and $c^{\ast}_i$ denote the individual probability distribution of different predicted classes and the ground-true class respectively. Here, $h_i$ and $o_i$ represent the estimated regression offsets of HBB and OBB. $h^{\ast}_i$ and $o^{\ast}_i$ denote the targeted object regression offsets, respectively. The hyper-parameters $\lambda_1-\lambda_5$ are the balance factors of different losses and for simplicity we set all to 1 in our experiments. $N_{cls}, N_{reg}, N^{\ast}_{cls}$, and $N^{\ast}_{reg}$ are the normalization factors to minimize the effects of different objects' scales when calculating losses which help to utilize an optimal training process.
In our experiments, the learning rate is initially set to $2\times 10^{-2}$ for $50k$ iterations, and progressively decreases to $10^{-3}$ and $10^{-4}$ for another $20k$ iterations. The batch size, momentum and the weight decay are set to $16$, $0.9$ and $0.0005$ respectively. The whole set of the filters were initialized from a regular distribution to the one of zero mean and $\sigma = 10^{-4}$.\\
To evaluate the performance of REFIPN for the estimation of the object orientations, we use the DOTA-2 dataset (plane and car classes). Table \ref{tab:2} outlines the mean test error. As the evaluation results show, REFIPN, in average by more than 18\% improvements, substantially outperforms the other state-of-the-art models. 
As shown in Fig. \ref{fig:6}, due to the lack of rotation equivariance, the standard CNN does not properly generalize for orientation changes. By utilizing augmentation (rotation), the error significantly decreases. This is the reason why the network has learnt to detect the rotated samples, leading to improvement of the learning capacity. Despite this turn-out, the proposed RConv has a satisfactory generalization ability across different orientations even without augmentation.

\begin{figure*}
  \centering
  \includegraphics[width=0.95\textwidth]{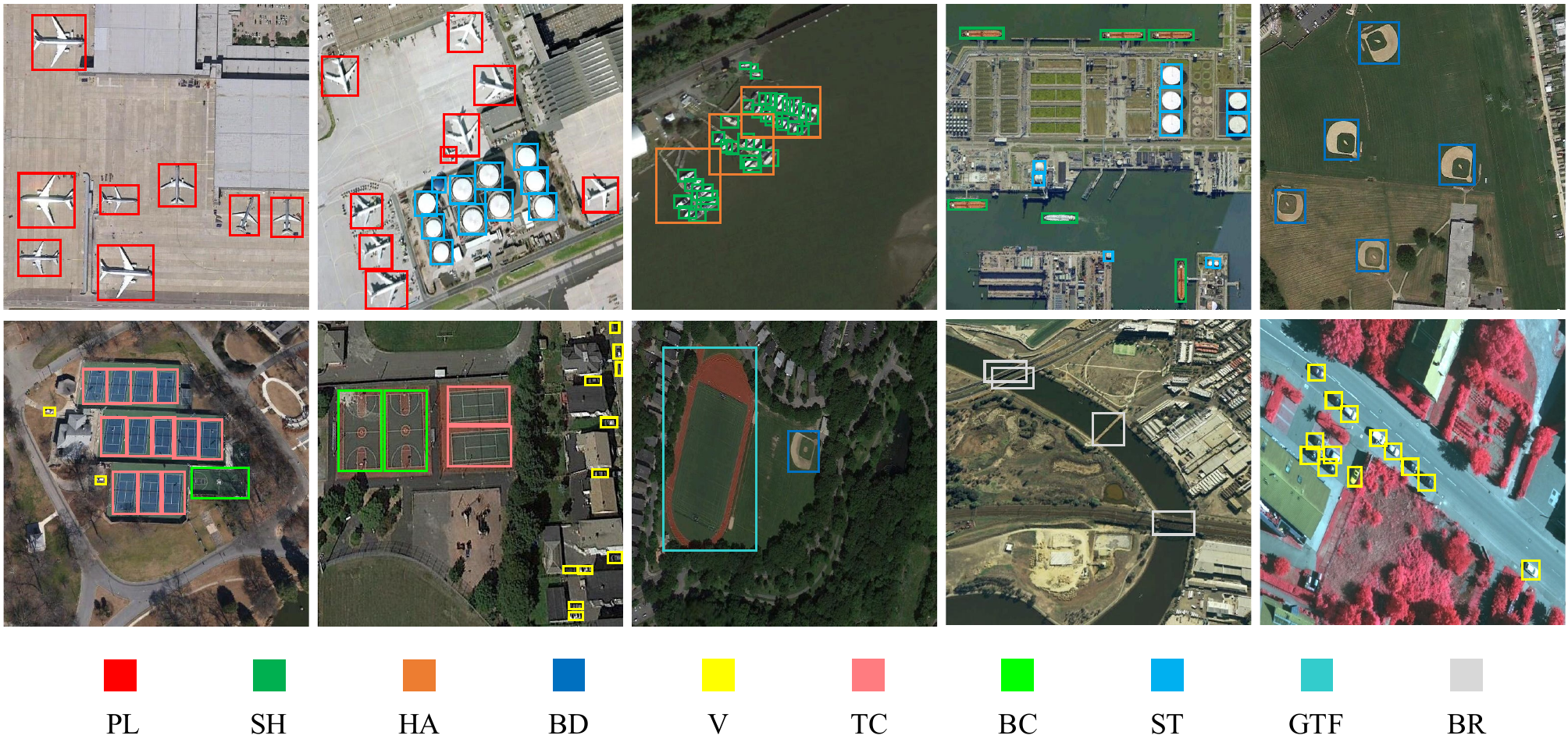}
\caption{Exemplar detection results on the test NWPU VHR-10 dataset. Plane (PL), Ship (SH), Harbor (HA), baseball diamond (BD), Vehicle (V), Tennis court (TC), Basketball court (BC), Storage tank (ST), Ground track field, (GTF), and Bridge (BR).}
\label{fig:12}
\end{figure*}
\subsection{Comparison with State-of-the-art Methods}
In this section, we compare the performance of our proposed REFIPN against that of the other state-of-the-art approaches. 

In Tables \ref{tab:3} and \ref{tab:5}, we respectively show the results of our detector in comparison with the other models on the DOTA and NWPU VHR-10 datasets. On the DOTA dataset, for the $400 \times 400$ pixel input image, the standard SSD attains a detection rate of $68.54$ mAP while operating at $134$ FPS. Among the current object detectors in RSIs, HSP \cite{xu2020hierarchical} and LR-CNN \cite{guo2020rotational} achieve the detection rates of $78.25$ and $76.59$ mAP while operating at $37$ and $65$ FPS, respectively. Our proposed REFIPN achieves a satisfactory trade-off between the precision of detection and speed with $79.54$ mAP while operating at $112$ FPS. Some sample detection results are shown in Fig. \ref{fig:7}. Our proposed detector on the NWPU VHR-10 dataset also obtains the state-of-the-art results and outperforms the other rotation equivariance methods. Some sample detection results on the test set of the NWPU VHR-10 dataset are shown in Fig. \ref{fig:12}. Our detector on the 800 pixel input image achieves 92.86\% mAP which shows the superiority of our model to the other approaches proposed for RSI object detection. Especially for plane, vehicle, and several other classes, our method achieves substantial performance improvement as reported in Table \ref{tab:5}.
This improvement is largely due to the following factors. 
\begin{itemize} 
\item[1)] The proposed RConv layer improves the learning capacity and satisfies self-consistency and invariance in absolute orientation estimation, which leads to more accurate object detection. 
\item[2)] By integrating the proposed LIPM into the standard SSD architecture, we build a feature pyramid network in which each scale of the image pyramid is featurized that boosts the discriminative ability of the detector. 
\item[3)] The feature fusion module can transform the central attention of the detector from the key parts to the whole objects. Therefore, more accurate detection can be achieved.
\end{itemize}
\begin{figure}
  \centering
  \includegraphics[width=0.47\textwidth]{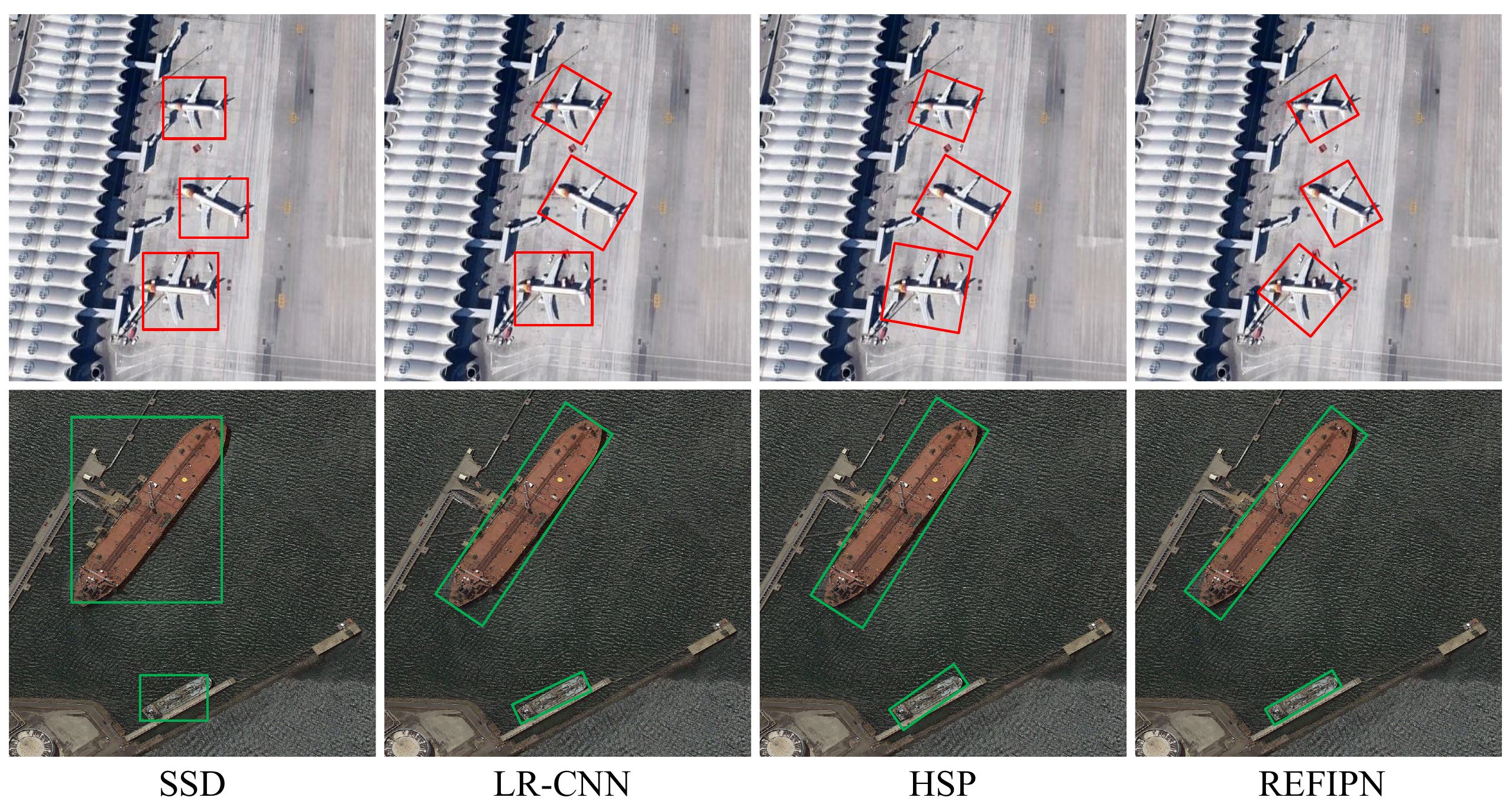}
\caption{Qualitative detection comparison by different models on the DOTA dataset. }
\label{fig:8}
\end{figure}
\begin{figure}
  \centering
  \includegraphics[width=0.47\textwidth]{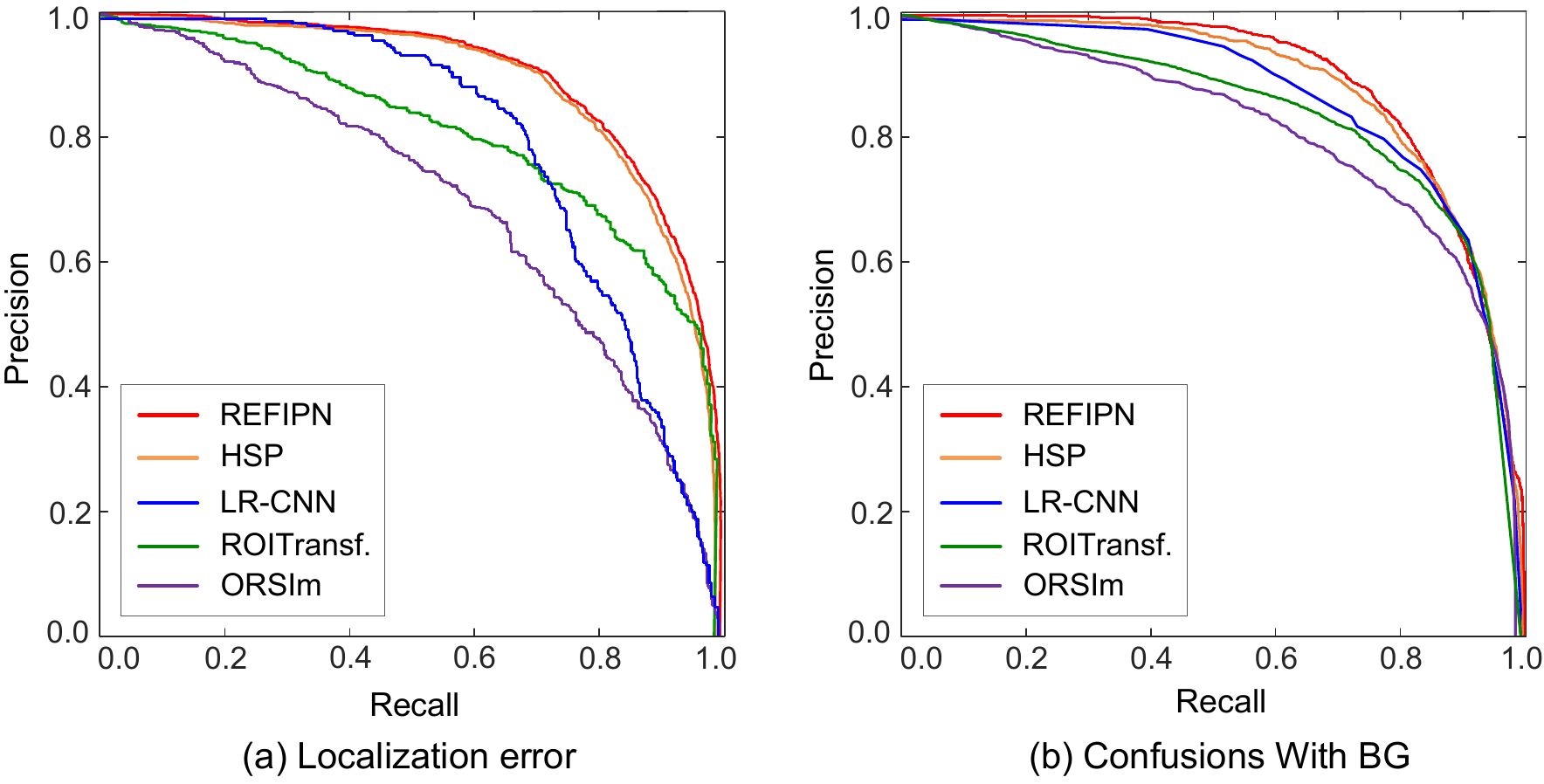}
\caption{Analysis of detectors performance on DOTA dataset. (a) Localization error curves. (b) Confusions with BG.}
\label{fig:ROC}
\end{figure}
\begin{figure}
  \centering
  \includegraphics[width=0.47\textwidth]{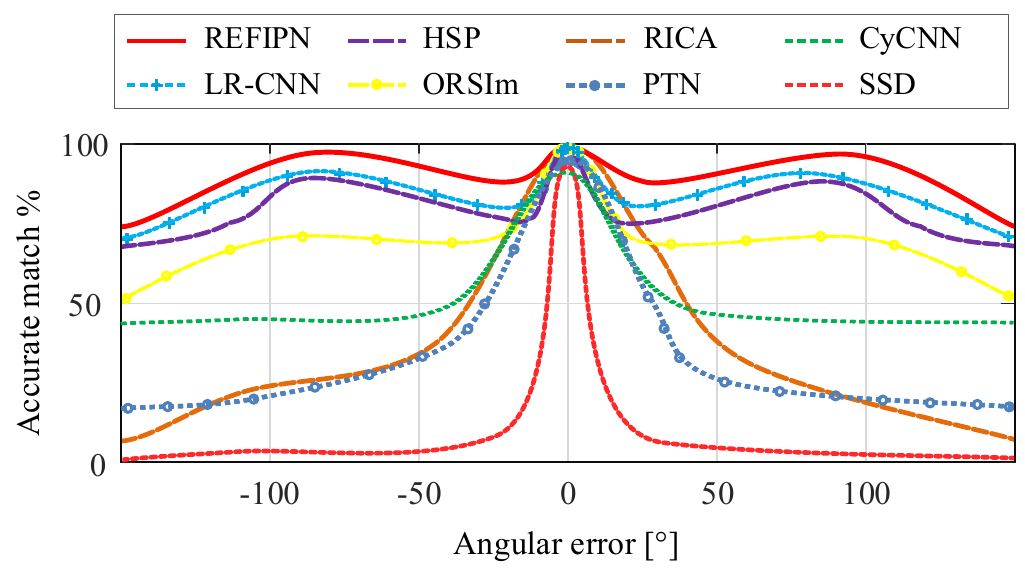}
\caption{Accuracy $vs.$ rotation. We observe REFIPN is significantly more robust than the state-of-the-arts for rotation estimation.}
\label{fig:9}
\end{figure}
To show the advantages of REFIPN as compared to LR-CNN \cite{guo2020rotational} and HSP \cite{xu2020hierarchical}, which show the second and third best performers in our experiments, where qualitative performance comparison of different methods on various scales and orientations of objects is conducted. As Fig. \ref{fig:8} shows, the other two methods are less satisfactory in detecting the objects in images, and the background is mis-detected as the foreground mostly due to the orientations of the objects. As the results show, in the other methods, the BBs are not well fit to the detected objects, however, our detector can stably produces precise results.
To evaluate the performance of our proposed model, in Fig. \ref{fig:ROC} we plot the mean localization error and confusions with BG curves over the DOTA dataset. As the results show the proposed model has higher localization and classification accuracy compared to the other baselines. This performance is due to structure and response of our model to the rotation of an object, constructing rotation-equivariant features. \\ 
Fig. \ref{fig:9} evaluates the performance of REFIPN and the other approaches with respect to object rotation. As the results show, our proposed model is significantly more robust against rotations as compared to the other methods. Our method substantially outperforms the other methods on the image with tiny angles and still have more than 85\% of accurate estimations for the rotations around $45^{\circ}$ while less than 70\% is achieved by LR-CNN \cite{guo2020rotational} and HSP \cite{xu2020hierarchical} and less than 50\% is achieved by PTN \cite{esteves2017polar} and CyCNN \cite{kim2020cycnn}.

\subsection{Ablation Study}
To evaluate the significance of each proposed module within the proposed framework, we perform a comprehensive
ablation study and show the results in Table V that reports the comparison, with respect to detection accuracy and speed of our proposed model against those of the baseline SSD on the DOTA dataset.
\begin{table}
\centering
  \caption{ABLATION RESULTS FOR HBB TASK ON DOTA TEST SET (800 PIXELS) WITH LIPM AND FFM AT VARIOUS LEVELS OF SSD ADOPTED RCONV.}
  \label{tab:4}
  \begin{tabular}{p{0.065\textwidth}|p{0.066\textwidth}p{0.026\textwidth}p{0.026\textwidth}p{0.026\textwidth}p{0.026\textwidth}p{0.026\textwidth}p{0.026\textwidth}}     \hline
    Add to & SSD+RConv& & & & &  & Ours \\     \hline
RConv 4 & & $\hfil \checkmark$ & $\hfil \checkmark$  &  $\hfil \checkmark$ & $\hfil \checkmark$ &   $\hfil \checkmark$ & $\hfil \checkmark$ \\
RConv 9  & & & $\hfil \checkmark$  & $\hfil \checkmark$ & $\hfil \checkmark$  & $\hfil \checkmark$ & $\hfil \checkmark$  \\
RConv 10  & & & & $\hfil \checkmark$  & $\hfil \checkmark$ & $\hfil \checkmark$ & $\hfil \checkmark$  \\
RConv 11 & & & & & $\hfil \checkmark$ & $\hfil \checkmark$ & $\hfil \checkmark$ \\
RConv 12 & & & & & & $\hfil \checkmark$ &  \\
with FFM   & & & & & & & $\hfil \checkmark$ \\ \hline
mAP & 72.46	&75.68&	76.42	&77.23&	78.97	& 79.02 & {80.43} \\
FPS &  66  	& 65 &	63	& 60 &	58	& 57 & {57} \\
FLOPS &  36B  	& 40B &	46B	& 50B &	57B	& 61B & {58B} \\
        \hline
\end{tabular}
\end{table}
\subsubsection{Rotation Equvariant Convolution}
\begin{figure}
  \centering
  \includegraphics[width=0.5\textwidth]{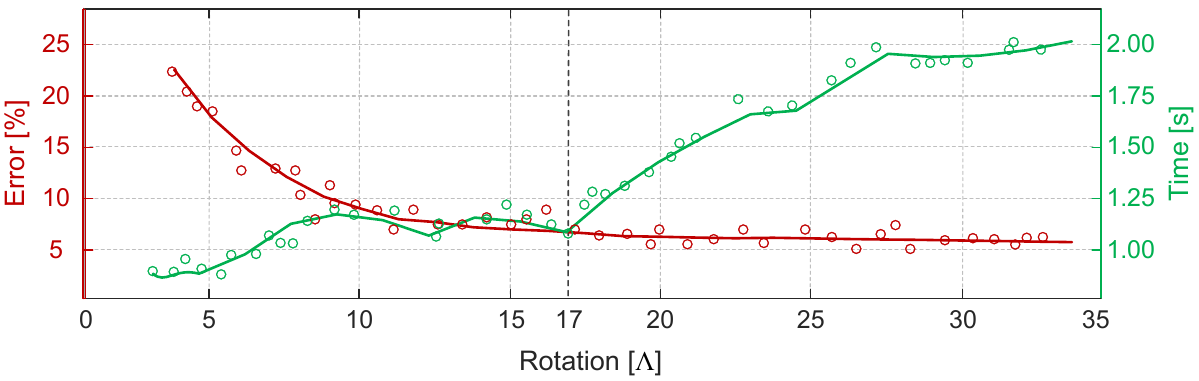}
\caption{Test error (left) vs. computation time (right) for different numbers of filters rotation.}
\label{fig:13}
\end{figure}

As shown in Tables \ref{tab:3} and \ref{tab:4}, the mAP improves from 69.23\% to 72.46\% on the $800$ pixel input image by using RConv in SSD. It can be observed that the standard CNN does not perform well for the orientation estimation. By adopting rotational augmentation, its error substantially decreases, however, for tiny angles (near to zero), it grows again. On the other hand, the RConv performs well enough on orientation estimation even without using augmentation. 
In Fig. \ref{fig:13}, we evaluate the sensitivity of REFIPN to $\Lambda$. To Investigate the sensitivity of REFIPN to the number of rotations $\Lambda$, we use $\Lambda = 24$ to train the model and test it for different rotation values. We have noticed for $\Lambda > 17$ there are small changes in the test error, however there is a significant increase in computation time. 
\subsubsection{Light-weight Image Pyramid and Feature Fusion Module}
To evaluate the effect of the proposed LIPM on the SSD, we conduct a set of experiments by continuously inserting layers at the LIPM and combine them with the SSD's layers. Table \ref{tab:4} reports each layer's detection results, speed and FLOPS. Large improvement (3.22\% mAP) is attained when we combined the feature image pyramid network with the RConv4. The performance of the detection is further improved in the following levels and when we use the feature fusion modules, resulting in detection accuracy of $80.43$ mAP. We evaluate our proposed model on various settings to design an effective and efficient architecture for our proposed model. As Table \ref{tab:4} shows, by adding one more layer to the LIPM and merge it with the matching layer of the SSD ($RConv12$), we achieve slight improvement in detection, however, there is a significant increase in computation cost.  
\begin{figure}
  \centering
  \includegraphics[width=0.49\textwidth]{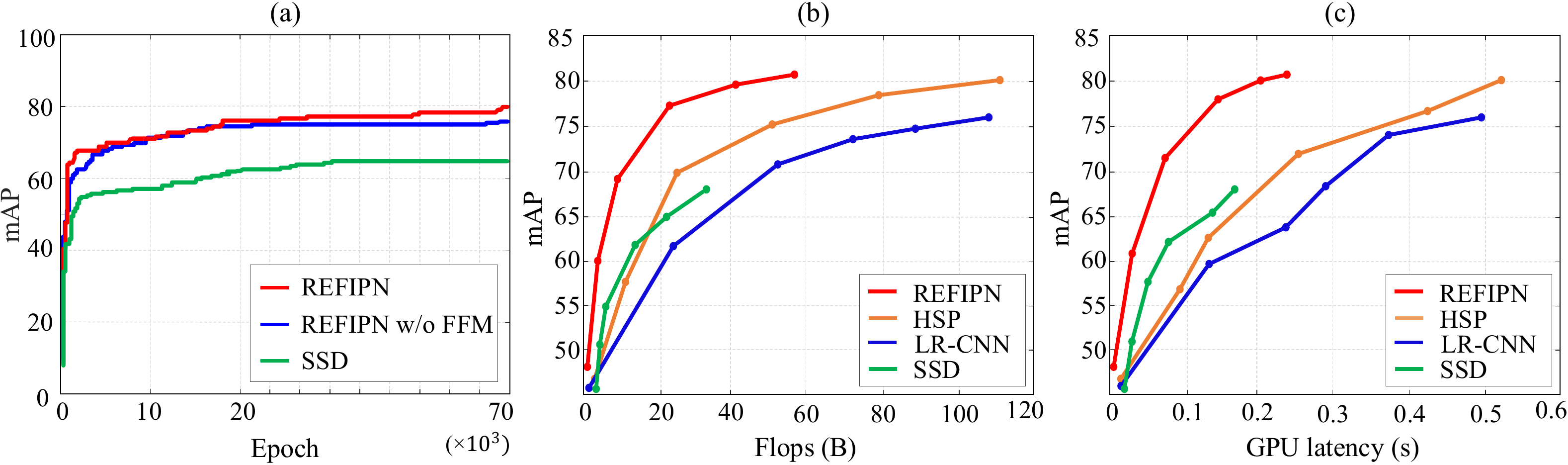}
 \caption{Evaluation of different methods on the DOTA dataset: (a) Performance evaluation of FFM. (b) and (c) Efficiency comparison of different methods. Performance is measured on the same machine equipped with a Tesla V100 GPU.}
\label{fig:10}
\end{figure}
As previously discussed, we propose a FFM that preserves the benefits of the normalized weights and combines features from both the current and previous layers. In Fig. \ref{fig:10}(a), we compare the performance of REFIPN with and without adopting FFM. By adopting the FFM, our proposed model achieves higher accuracy with better learning abilities. In Fig. \ref{fig:10} (b) and (c) we illustrate the FLOPS-accuracy and GPU latencey-accuracy curves for the models, where REFIPN achieves higher accuracy in shorter time with much fewer FLOPS against the other state-of-the-art models. As the results illustrate, our detector attains better efficiency as compared to the other models, which signifies the advantages of RConv and the joint scale-wise learning.

\begin{figure}
  \centering
  \includegraphics[width=0.45\textwidth]{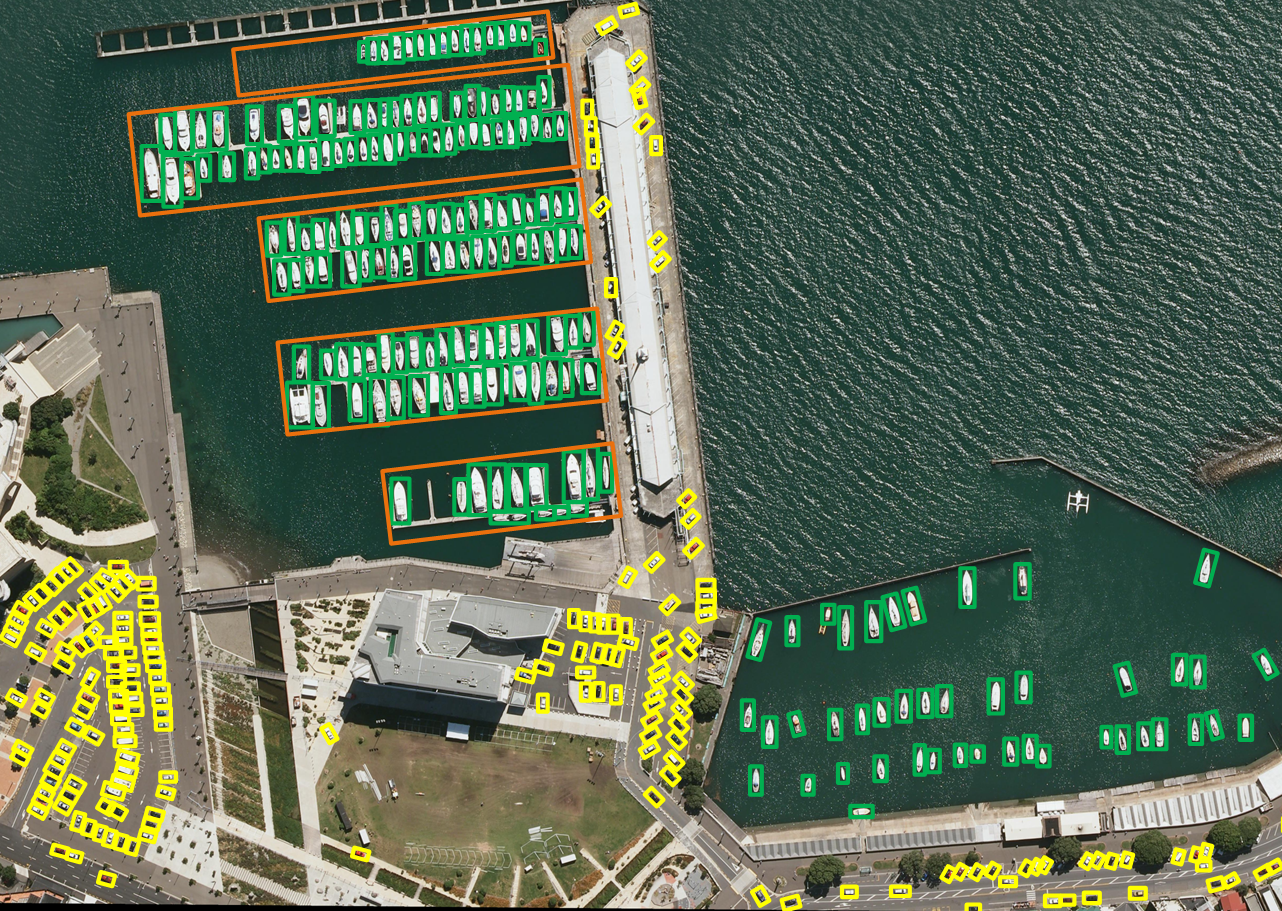}
 \caption{Detection results on a large-scale RSI. Green, orange and yellow respectively show ships, harbors, and cars.}
\label{fig:11}
\end{figure}

In Fig. \ref{fig:11}, we reveal our model's performance on a large scale of RSI ($1600 \times 1600$ pixels). It is observed that the pre-trained REFIPN performs consistently on different image sizes and conditions.

\section{Conclusion}
\label{sec:5}
In this paper, we introduced an accurate and efficient object detection architecture in RSIs, called REFIPN, by considering rotation equivariant within CNNs operations and using the proposed feature image pyramid network to extract semantic features in a wide range of scales. This was achieved by implementing several filters to handle various orientations and acquiring a vector field feature map to use the highest activation with respect to magnitudes and angles in the estimations. Extensive experiments on orientation estimation and the detection results showed that our proposed REFIPN model performs better than the other approaches. The results justify that considering the predominant orientations is effective in tackling a wide range of challenging problems. In our future work, we will aim to develop a more efficient feature learning algorithm to have a scale adaptation ability to further improve detection accuracy of small-sized objects in aerial scenes.

\footnotesize{
\bibliographystyle{IEEEtran}
\bibliography{IEEEabrv,IEEEexample}
}


\end{document}